%% file: main.tex
\documentclass[acmtog]{acmart}
\usepackage{xcolor}
\usepackage{graphicx}
\usepackage{caption}
\usepackage{subcaption}
\usepackage{siunitx}
\usepackage{multirow}
\usepackage{makecell}

\usepackage{rotating}
\usepackage{amsfonts}
\usepackage{placeins}
\usepackage{graphbox}

\setcopyright{none}
\usepackage{enumitem}

\copyrightyear{2025}
\acmYear{2025}
\setcopyright{cc}
\setcctype{by-nc}
\acmConference[SIGGRAPH Conference Papers '25]{Special Interest Group on Computer Graphics and Interactive Techniques Conference Conference Papers }{August 10--14, 2025}{Vancouver, BC, Canada}
\acmBooktitle{Special Interest Group on Computer Graphics and Interactive Techniques Conference Conference Papers (SIGGRAPH Conference Papers '25), August 10--14, 2025, Vancouver, BC, Canada}
\acmDOI{10.1145/3721238.3730592}
\acmISBN{979-8-4007-1540-2/2025/08}

\begin{document}

\title{Uncertainty for SVBRDF Acquisition using Frequency Analysis}

\author{Ruben Wiersma}
\email{rubenwiersma@gmail.com}
\affiliation{%
 \institution{ETH Zurich,}
 \institution{Delft University of Technology,}
 \institution{Adobe Research}
 \country{Switzerland}
 }
\author{Julien Philip}
\affiliation{%
 \institution{Adobe Research,}
 \institution{Netflix Eyeline Studios}
 \country{UK}
 }
\author{Miloš Hašan}
\affiliation{%
 \institution{Adobe Research}
 \country{USA}
 }
 \author{Krishna Mullia}
\affiliation{%
 \institution{Adobe Research}
 \country{USA}
 }
 \author{Fujun Luan}
\affiliation{%
 \institution{Adobe Research}
 \country{USA}
 }
 \author{Elmar Eisemann}
\affiliation{%
 \institution{Delft University of Technology}
 \country{The Netherlands}
 }
 \author{Valentin Deschaintre}
 \email{deschain@adobe.com}
\affiliation{%
 \institution{Adobe Research}
 \country{UK}
 }

\begin{abstract}
  This paper aims to quantify uncertainty for SVBRDF acquisition in multi-view captures. Under uncontrolled illumination and unstructured viewpoints, there is no guarantee that the observations contain enough information to reconstruct the appearance properties of a captured object. We study this ambiguity, or uncertainty, using entropy and accelerate the analysis by using the frequency domain, rather than the domain of incoming and outgoing viewing angles. The result is a method that computes a map of uncertainty over an entire object within a millisecond. We find that the frequency model allows us to recover SVBRDF parameters with competitive performance, that the accelerated entropy computation matches results with a physically-based path tracer, and that there is a positive correlation between error and uncertainty. We then show that the uncertainty map can be applied to improve SVBRDF acquisition using capture guidance, sharing information on the surface, and using a diffusion model to inpaint uncertain regions.
  Our code is available at \url{https://github.com/rubenwiersma/svbrdf_uncertainty}.
\end{abstract}

\ccsdesc[500]{Computing methodologies~Reflectance modeling}
\ccsdesc[500]{Computing methodologies~3D imaging}
\ccsdesc[500]{Computing methodologies~Appearance and texture representations}

\keywords{Uncertainty, Inverse rendering}

\begin{teaserfigure}
 \includegraphics[width=\textwidth]{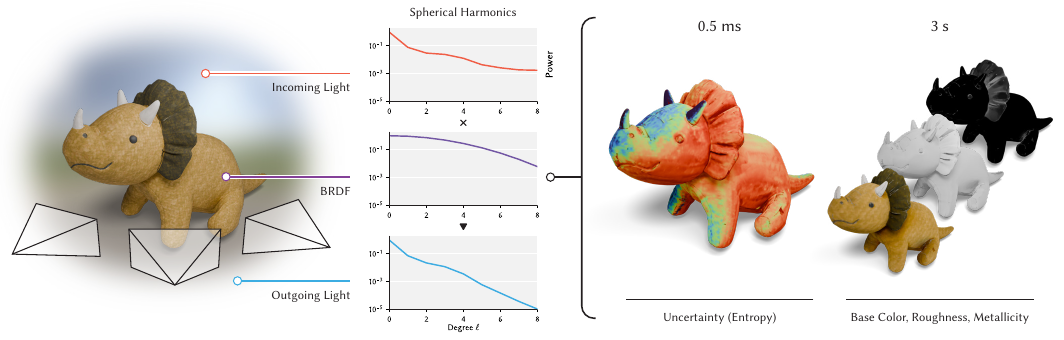}
\centering
\caption{We quantify uncertainty for SVBRDF acquisition from multi-view captures (left) using entropy (right). We significantly accelerate the, otherwise heavy, computation in the frequency domain (spherical harmonics power spectrum, center), leading to a practical, efficient method.}
 \label{fig:teaser}
 \Description[A visual summary showing a stuffed dinosaur model under lighting conditions, spherical harmonics graphs, uncertainty mapping, and material property predictions (base color, roughness, metallicity).]{
 The figure illustrates a pipeline for estimating uncertainty and spatially-varying material properties of a plush dinosaur toy. On the left, a rendered model shows the input: incoming lighting, the BRDF, and outgoing lighting from multi-view captures. Three plots in the center display spherical harmonics power spectra for incoming lighting, BRDF, and outgoing lighting. Right side: first, a colored uncertainty map (entropy) over the dinosaur surface is shown, computed within 0.5 milliseconds. Then, three images show estimated base color, roughness, and metallicity maps, computed in about 3 seconds.
 }
 \end{teaserfigure}

\maketitle

\input{sections/1_introduction}
\input{sections/2_related}
\input{sections/3_method}

\input{sections/4_experiments}
\input{sections/5_conclusion}
\input{sections/6_acknowledgements}

\bibliographystyle{ACM-Reference-Format}
\bibliography{references}

\appendix
\input{sections/A_appendix}

\end{document}

%% file: sections/1_introduction.tex
\section{Introduction}
Recovering realistic appearance properties for digital objects is inherently challenging. We propose to estimate the uncertainty of the process, given a set of views of an object and information about the incident lighting.
This enables applications such as viewpoint planning and information sharing for improved acquisition. 

Appearance is typically modeled as a six-dimensional function, depending on the incoming and outgoing light directions and locations on the surface.
Accurately reconstructing this high-dimensional function requires many samples and, ideally, control of both lighting and camera positions.
Such capture requires complex and expensive hardware, such as light stages, and is particularly difficult for on-site capture. 
It is more convenient to \textit{passively} photograph the object from a set of viewpoints, without control of the lighting.

\begin{figure}[t]
    \centering
    \includegraphics[width=\linewidth]{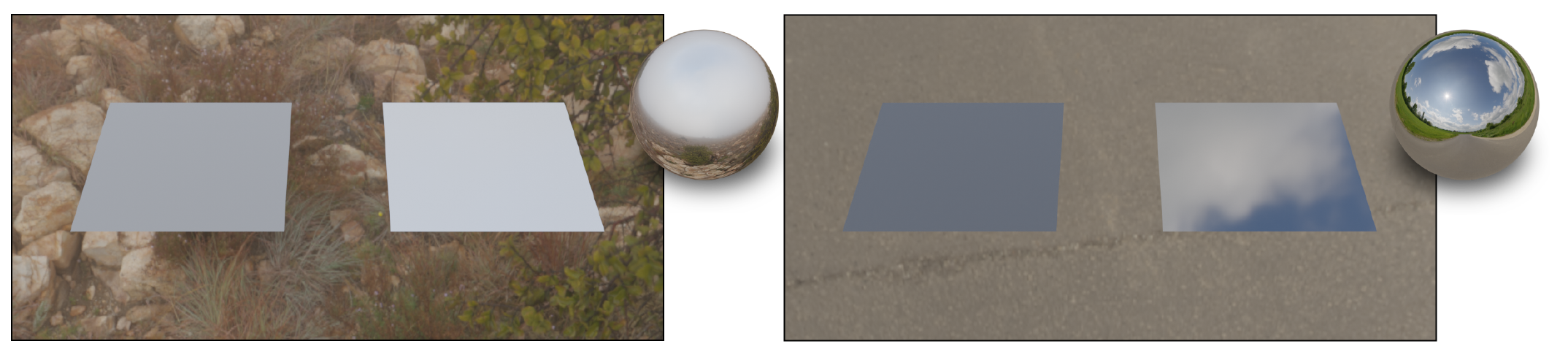}
    \caption{Passive capture can lead to ambiguity about the material of an object. Which plane is glossy and which is matte? While it is hard to see in the left scene, the materials are the same in both scenes (left matte, right glossy).}
    \label{fig:problemstatement}
    \Description[Two side-by-side examples showing material ambiguity: identical flat surfaces under different outdoor lighting conditions appear differently, with environment maps shown as spheres.]{
    The figure highlights material ambiguity caused by illumination. On the left, two identical gray planar surfaces are shown under an overcast environment, appearing almost indistinguishable. A spherical environment map of the overcast scene is included. On the right, the same surfaces are placed in a different outdoor environment (an asphalt road under a bright sky). Here, one surface appears significantly more mirror-like due to lighting effects. A corresponding spherical environment map of the road scene is also shown. The figure illustrates how the same material can look very different under varying lighting, complicating material estimation.
    }
\end{figure}

Passive captures often lead to ambiguity about the material of an object, as shown in \autoref{fig:problemstatement}. 
Depending on the lighting and viewpoint, it may be hard to distinguish matte from glossy materials.
On complex, curved shapes, such ambiguities can arise, e.g., from occlusion, shadowing, and missing viewpoints.
In this work, our main goal is to estimate the ambiguity, or \textit{uncertainty}, when recovering SVBRDF parameters.
The result is a map on the surface, quantifying how certain we are about the parameter estimates (\autoref{fig:teaser}).
We demonstrate that this information can improve SVBRDF recovery through capture guidance, information sharing, and inpainting of uncertain areas.
We also show that our contributions can be used for fast recovery of SVBRDF parameters. 

To quantify uncertainty, we use entropy.
Entropy measures how `spread out' a probability distribution is: in the overcast setting (left of \autoref{fig:problemstatement}), the planes being glossy or matte is equally likely.
The probability distribution is uniform, leading to high entropy. In the setting on the right, it is very unlikely that the plane on the right is matte.
The probability distribution is concentrated around the glossy material, resulting in low entropy.
To compute entropy, we require knowledge of the posterior probability distribution over the entire space of possible parameters.
This requires rendering the scene from every input viewpoint for all possible parameter combinations.
Prior approaches for estimating uncertainty handle this by sparsely sampling the parameter space. For example, \citet{lensch} and \citet{goli2023} use the inverse of the Hessian around the optimized parameters as a proxy for uncertainty. This approach, however, is local in parameter space and requires running an optimization first. \citet{zhou2024estimating} sample the full parameter space using stochastic particle-optimization sampling (SPOS)~\cite{zhang2020spos}.
While this approach considers the full parameter space, it is slow to compute.

Rather than focusing on reducing the number of samples, we accelerate the computation of the probability for each parameter combination using frequency-domain analysis~\cite{Ramamoorthi_Hanrahan_2001}.
The resulting approach is lightweight and highly parallelizable, making entropy as a measure of uncertainty practical for SVBRDF recovery.
The frequency-domain analysis considers reflection functions as convolutions of the BRDF over the incoming light, easily computed in the frequency domain.
We show that the entire reflection function can be analyzed within the frequency domain, given careful approximations.
This eliminates repeated transforms to the angular domain.
Second, we show that the analysis of uncertainty within the frequency domain can be accelerated by only considering the power spectra of spherical harmonics.
We also find that uncertainty is determined by two parameters, the specular coefficient and roughness, further reducing the complexity of the problem.
To be practical, these contributions require a spherical-harmonics transform for sparse and irregular samples of the incoming and outgoing radiance, as these are sampled through unstructured photographs. We therefore propose a robust spherical-harmonics transform that uses least-squares fitting on sparse and irregular points, with regularization specialized for natural lighting.

We validate our method by comparing our entropy estimation to entropy computed with a state-of-the-art path tracer~\cite{jakob2022mitsuba3}. We find that our approach correlates strongly ($\rho=0.9$) and runs in a fraction of the time ($1/700\,000$). We also evaluate the relationship between entropy and reconstruction error and show three applications of our uncertainty evaluation: capture guidance, information sharing between certain and uncertain areas on the surface, and inpainting regions with high uncertainty using a diffusion model. Finally, we demonstrate that our framework can be used to optimize SVBRDF parameters with state-of-the-art relighting results and a $10$x speedup and use these results to initialize other approaches.

In summary, we propose to compute uncertainty for the common passive capture setup, through the following contributions:
\begin{itemize}[noitemsep,topsep=2pt,parsep=0pt,partopsep=0pt]
    \item Uncertainty for SVBRDF acquisition through entropy.
    \item A practical and fast entropy computation, following from an analysis in the frequency domain and power spectrum simplification.
    \item Demonstration of applications for uncertainty.
\end{itemize}

%% file: sections/2_related.tex
\section{Related Work}
We provide a detailed overview of related work in uncertainty estimation for (SV)BRDF recovery and a brief analysis of how approaches for (SV)RBDF recovery deal with ambiguity.

\subsection{Uncertainty Estimation}
Similar to us, prior works study uncertainty from a Bayesian perspective,  analyzing the posterior probability distribution over BRDF parameters.
Most works sample parameters sparsely, to avoid costly rendering operations.
We distinguish approaches that sample the parameter space locally (Laplace's approximation) or globally.
Laplace's approximation assumes the posterior follows a Gaussian distribution centered around the optimal parameters and uses the inverse of the Hessian as a covariance matrix, representing uncertainty.
\citet{lensch} use this approach to select viewpoints that minimize uncertainty.
In another work, \citet{lensch03} use the covariance matrix to optimally split clusters of materials.
Recently, \citet{goli2023} use Laplace's approximation to quantify uncertainty for NeRFs~\cite{mildenhall2020nerf}.
They also show a correlation between uncertainty and absolute error.
This information is used to remove spurious geometry.
In our work, we approximate the rendering equation, which allows us to sample the posterior probability distribution globally along a dense grid. We directly use entropy to measure uncertainty, rather than the covariance matrix.

\begin{figure*}
  \includegraphics[width=\textwidth]{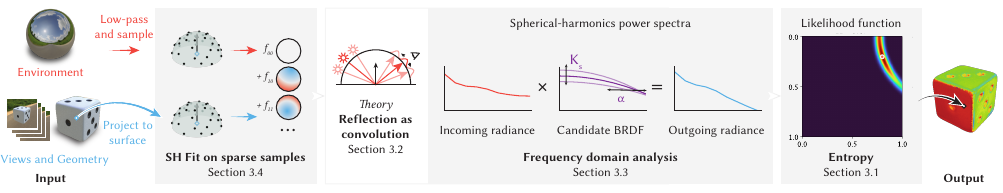}
 \centering
    \caption{An overview of our pipeline showing input, output, and the proposed algorithm. The input to our approach is a set of photographs of an object from multiple viewpoints and the associated camera extrinsics and intrinsics. We also provide the input lighting as an environment map and object geometry. We first estimate spherical harmonic coefficients on both the incoming and outgoing radiance and then estimate the effect of different BRDF filters within the power spectrum. This is used to compute a measure of uncertainty for the predicted parameters of the acquisition.}
  \label{fig:overview}
  \Description[The figure shows an overview of our pipeline with input, output, and the proposed algorithm.]{The figure shows an overview of our pipeline with input, output, and the proposed algorithm. Left: The input to our approach is a set of photographs of an object from multiple viewpoints and the associated camera extrinsics and intrinsics. We also provide the input lighting as an environment map and object geometry. Next, we estimate spherical harmonic coefficients on both the incoming and outgoing radiance and then estimate the effect of different BRDF filters within the power spectrum. This is used to compute a measure of uncertainty for the predicted parameters of the acquisition.}
\end{figure*}

\citet{zhou2024estimating} sample the full parameter space using stochastic particle-optimization sampling (SPOS)~\cite{zhang2020spos}.
SPOS optimizes samples through particle-optimization to be distributed according to a target function that resembles the posterior.
Then, the variance of these samples is used as a proxy for uncertainty. These contributions are orthogonal to our work.
Because we evaluate the rendering equation in the frequency domain, our approach takes roughly one millisecond to compute uncertainty, while \citet{zhou2024estimating} report ``a couple to 20 minutes'' on comparable hardware.

\citet{umat} used Monte-Carlo dropout to estimate uncertainty in the context of single-view material estimation, taking inspiration from Bayesian methods (\cite{gal2016dropout}). Our method is designed for multi-view captures and measures uncertainty in the input captures, rather than the uncertainty of a predictive model.

\subsection{Frequency-Based Light Transport}
Our work builds on the idea that BRDFs can be expressed as low-pass filters in the frequency domain~\cite{Durand2005, Ramamoorthi_Hanrahan_2001}.
This concept is widely adopted in interactive rendering, e.g., \cite{bagher2013interactive}. We apply it in the context of material acquisition, where it has been used for controlled illumination~\cite{Ghosh_Achutha_Heidrich_O’Toole_2007, aittala2013practical}, controlling the frequency of light patterns to estimate the BRDF filter parameter through deconvolution of the reflected light.
We do not assume control of the light and operate in the spherical-harmonics frequency domain rather than the Fourier domain \cite{aittala2013practical} or custom basis functions \cite{Ghosh_Achutha_Heidrich_O’Toole_2007}.
Our analysis therefore works with arbitrary natural lighting environments.

Our work is closest in spirit to that of \citet{Ramamoorthi_Hanrahan_2001}. 
\citeauthor{Ramamoorthi_Hanrahan_2001}'s analysis allows one to conclude whether the BRDF recovery is well-posed or not, given prior knowledge of the width of the normal distribution function.
Our uncertainty estimation is easier to apply as it gives a more precise and continuous answer to the question of uncertainty and does not assume prior knowledge of the material.
Second, our approach works directly in the frequency domain and supports arbitrary light setups by using a robust spherical-harmonics transform.
Finally, we improve the BRDF model to include shadowing and masking and demonstrate that our improvements to the method result in competitive performance in (SV)BRDF capture.

\subsection{(SV)BRDF Recovery}
Our method operates in the context of (SV)BRDF recovery using optimization-based methods~\cite{munkberg2022extracting, Loubet2019Reparameterizing, nimierdavid2021material, Vicini2022sdf}.
These methods optimize BRDF parameters and, optionally, geometry and lighting, to reproduce the appearance of a set of input views. 
Existing approaches try to reduce the ambiguity in their capture, which can lead to suboptimal convergence.
A direct way to reduce ambiguity is to use many photographs and controlled lighting~\cite{aittala2013practical,MobileSVBRDF:SIGA:2018,Dupuy2018Adaptive} or object orientations~\cite{dong2014appearance}.
Some rely on specialized hardware to capture polarimetric information~\cite{Ellipsometry:SIG:2022}.
Others use prior knowledge to compensate for limited information, such as stationarity of the captured materials~\cite{aittala2015two, aittala2016one, xu2016minimal, henzler2021neuralmaterial}.

Another way to reduce ambiguity is to use data-based priors. This allows the capture of (SV)BRDFs from as little as one image~\cite{li2017modeling, MRRKB:2022:MaterIA, vecchio2023controlmat, Shi2020:ToG, deschaintre2018single, deschaintre2019flexible, deschaintre20, join23, xilong2021ASSE, Zhou2022look-ahead, zhou2023semi,luo:2024,Guo21-HA}. 
For material acquisition on curved surfaces, approaches have been proposed to extract material parameters from single, multiple flash, or multi-focal photographs~\cite{li2018learning, deschaintre2021deep, boss2020two, Fan23}. More recently, diffusion priors have been used to complete the missing lighting information in the context of acquisition~\cite{lyu2023dpi}.

Our method is intended to work in conjunction with general material acquisition approaches.
In our applications, we show several ways how our uncertainty analysis can be used to improve acquisition, both during optimization (e.g., by sharing information) and by using data-based priors (e.g., through diffusion model inpainting).

%% file: sections/3_method.tex
\section{Method}
Our goal is to evaluate uncertainty in material reconstruction during a `passive' capture. The result is a mapping on the surface that denotes where the signal is insufficient to guarantee an accurate reconstruction. The input to our method is a set of views from multiple viewpoints and we assume that the camera extrinsics, intrinsics, object geometry and HDR environmental lighting are known but not controlled.

\autoref{fig:overview} provides an overview of our method. In this section, we work our way back from the desired output, entropy, to a practical and efficient implementation. Entropy requires evaluating the reconstruction error of many material parameters for the entire surface, which we accelerate by building on the signal processing framework proposed by \citet{Ramamoorthi_Hanrahan_2001} (frequency domain analysis). We analyze the uncertainty problem in the frequency domain and find that, given some approximations, we can evaluate entropy entirely using the power spectrum of the spherical harmonics representation of the incoming and outgoing radiance. Finally, we show how to robustly transform sparse and irregular radiance samples in the angular domain to spherical harmonics.

\subsection{Entropy and Uncertainty}
\label{sec:method_uncertainty}

Entropy, in the context of information theory, measures the uncertainty about an unknown variable $x$ over the possible states $X$~\cite{shannon1948mathematical}. Given a probability distribution $p(x)$, entropy is defined in the discrete setting as
\begin{equation}
    H = -\sum_{x \in X} p(x)\log{p(x)}.
\end{equation}
Intuitively, entropy measures the `spread' of the probability distribution $p(x)$. For example, a uniform distribution has maximum entropy and a narrow, concentrated distribution has low entropy.
We are interested in measuring the uncertainty over the continuous parameter combinations $\boldsymbol\psi$ for a BRDF at each point on the surface (e.g., roughness, metallicity, base color), given the observed outgoing radiance $B$ and incoming radiance $L$.
For simplicity of exposition, we will consider $n$ evenly spaced discrete parameter choices $\psi$. This discretization can be adapted to incorporate perceptual non-linearity\footnote{Because we use the roughness parameterization from \citet{burley2012physically} (Eq. 11 in the Supplement), perceptual nonlinearity is already partly accounted for (p.15 in \citet{burley2012physically}).}. Computing continuous entropy also requires discretization, because an analytical solution is not available for the integral, and results in the same formulation for entropy we derive here.

First, we need to know the \emph{likelihood} that the parameters $\psi$ explain the observations $B$, given $L$, $p(\psi | B, L)$. This likelihood can be derived using Bayes' rule from $p(B|\psi, L)$, the \emph{probability} that we observe radiance $B$, given the parameters $\psi$ and incoming light $L$
\begin{equation}
    p(\psi|B, L) \propto p(B|\psi, L) p(\psi).
\end{equation}
In our experiments, we assume that $p(\psi)$ is uniform, but one could include priors on materials based on real-world data without affecting the remainder of this discussion. If our measurements and models were perfect, $p(B |\psi, L)$, would be $1$ for the radiance observation predicted by our BRDF model and $0$ otherwise. However, we encounter measurement noise and our renders are approximate. If we assume this error follows a Gaussian distribution with variance $\sigma^2$, we can model the associated probability distribution as
\begin{equation}
    p(B | \psi, L) = \frac{1}{\sigma \sqrt{2\pi}} \exp\left[-\frac{d(B, r(\psi, L))}{2\sigma^2}\right],
    \label{eq:posterior}
\end{equation}
where $d$ is the distance function between the renders $r(\psi, L)$ and the observed radiance $B$. Typically, $d$ is the $L1$- or $L2$ norm over the difference between the set of views and their corresponding renders for every viewpoint. To compute entropy, we need to normalize $p(\psi | B, L)$ to be a proper probability distribution, since, as a likelihood, it does not integrate to one:
\begin{equation}
    \bar{p}(\psi | B, L) = \frac{p(\psi | B, L)}{\sum_{\psi \in \Psi} p(\psi | B, L)}.
\label{eq:normalizedlikelihood}
\end{equation}
Finally, we would like the maximum entropy to be one for interpretability. Therefore, we divide entropy by $\log(n)$, the maximum entropy corresponding to a uniform distribution
\begin{equation}
    H = -\frac{1}{\log(n)}\sum_{\psi \in \Psi} \bar{p}(\psi | B, L) \log\bar{p}(\psi | B, L).\\
\label{eq:entropy}
\end{equation}

\begin{figure}[b]
    \centering
    \includegraphics[width=\columnwidth]{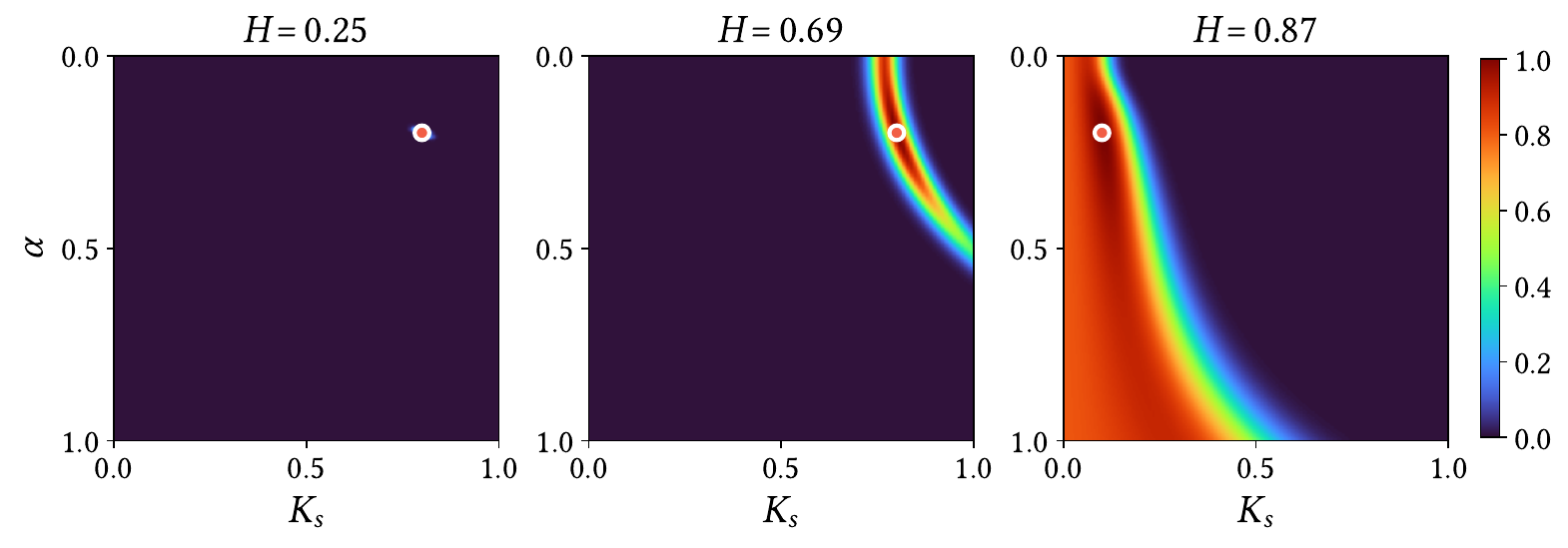}
    \caption{Examples of the likelihood and entropy $H$ for several parameter/lighting combinations. Left: An ideal situation, where the lighting is a dirac delta. Center: A situation where the lighting is too low-frequency to recover a good $\alpha$ value. Right: A situation where the incoming radiance is ideal, but the specular component is too low to get a proper recovery for $\alpha$.}
    \label{fig:entropy}
    \Description[The figure shows three plots with the likelihood function over alpha and K specular for different BRDF estimation problems.]{The figure shows three plots with the likelihood function over alpha and K specular for different BRDF estimation problems. The first plot has low entropy (0.25) and is highly concentrated around the optimum. The second plot has medium entropy (0.69) and shows a narrow sliver of likelihood around the ground-truth parameters. The third plot has high entropy (0.87) and is spread out over the entire range of alpha, around low K specular values.}
\end{figure}

\autoref{fig:entropy} shows three examples of likelihood over the roughness, $\alpha$, and the specular coefficient $K_s$, given different lighting conditions and ground-truth parameters (marked by a red dot). The heading $H$ denotes the corresponding entropy. An important observation is that these distributions do not follow a Gaussian distribution, which is the assumed distribution in Laplace's approximation. This demonstrates the importance of sampling the full parameter space, rather than a local approximation.

\subsection{Reflection as a Convolution}
\label{sec:reflection_as_convolution}
To compute entropy, we must evaluate the rendering function $r$ for the entire parameter space and for all viewpoints. 
For example, evaluating \autoref{eq:entropy} at $16$ intervals per parameter ($n=4096$) for $60$ viewpoints requires over 45 minutes with Mitsuba, a highly optimized path tracer.
To make our approach practical, we approximate the rendering equation using spherical harmonics, based on the signal processing framework for inverse rendering by~\citet{Ramamoorthi_Hanrahan_2001}.
\citeauthor{Ramamoorthi_Hanrahan_2001} show that the reflection function can be approximated with a spherical convolution of the BRDF over the incoming radiance.
Assuming an isotropic microfacet~\citet{torrancesparrow1967} BRDF, combined with a Lambertian term, they derive the following approximation for outgoing radiance $B$ at point $p$ in the view direction $\omega_o$
\begin{equation}
B(p, \omega_o)\approx K_d E(p) + K_s F(\theta_o)\left[S_\alpha \ast L(p) \right]_{\omega_o},
\label{eq:reflection_torrance_sparrow}
\end{equation}
where $K_d$ and $K_s$ are diffuse and specular terms; $E(p)$ is the irradiance integrated over the hemisphere; $F(\theta_o)$ is a simplified Fresnel term, which only depends on the outgoing direction; $L(p)$ is the incoming radiance; and $S_\alpha$ is a filter parametrized by the normal distribution width, $\alpha$. This filter is derived from the normal distribution function of the surface. The $\ast$ operator represents convolution. We provide a derivation of this approximation in the Supplement and detail there how to integrate the Shadowing and Masking terms~\cite{pharr2023physically} into the convolution perspective, which improves accuracy for high-roughness materials. The microfacet model that underlies this reflection model is still used in state-of-the-art inverse rendering research \cite{munkberg2022extracting,hasselgren2022shape,sun2023neural} using the parameters from the Disney Principled BRDF~\cite{burley2012physically}. In the Supplement, we show how to map the Principled BRDF parameters to our model and in our experiments, we validate that the model performs on par with a state-of-the-art path tracer for inverse rendering.

\subsection{Frequency Domain Analysis}
\label{sec:method_lightweight_objective}
Spherical harmonics-based rendering already significantly accelerates the evaluation of our uncertainty metric. We propose to go one step further for uncertainty estimation. The implementation of the rendering equation described in \autoref{eq:reflection_torrance_sparrow} computes convolution using spherical harmonics, but returns to the angular domain to evaluate radiance samples of $B$. Instead, we propose to evaluate the error function $d$ within the frequency domain, by first transforming the outgoing radiance $B$ to the frequency domain. This yields a significant speedup, as we only need to transform the outgoing radiance to the frequency domain once, rather than computing the inverse transform for every parameter combination. 
Moreover, we can evaluate the loss in the low-dimensional power spectrum, which allows us to evaluate all parameter combinations in parallel. The resulting speedup is of multiple orders of magnitude. This comes at the cost of some approximations, which we detail in this section. We show how to transform the outgoing radiance to the frequency domain in the next section.

The first required approximation is to omit the Fresnel, shadowing, and masking terms (detailed in the Supplement). These terms are applied as a pointwise multiplication in the angular domain, which is difficult to compute in the frequency domain. In Section~\ref{sec:quality_entropy}, we investigate the impact of this approximation and find that it is acceptable for uncertainty estimation. We can now rewrite \autoref{eq:reflection_torrance_sparrow} fully in the frequency domain, where $L_{\ell m}$ denotes the spherical harmonic coefficients for the incoming radiance and $B^\psi_{\ell m}$ the coefficients for the predicted outgoing radiance, given parameters $\psi = (K_d, K_s, \alpha)$
\begin{equation}
    B^\psi_{\ell m} = \begin{cases}
        {\pi}^{-\frac 1 2} K_d E(p) + K_s L_{00}, & \text{if $\ell=0$},\\
        K_s e^{-(\alpha \ell)^2}L_{\ell m}, & \text{otherwise}.
    \end{cases}
    \label{eq:specular}
\end{equation}
From this expression, we make two observations: (1) We cannot recover the ratio between $K_d$ and $K_s$ from degree $\ell=0$ alone, as we have two unknowns and only one coefficient. (2) $\alpha$ has no effect on degree $\ell=0$.
It follows that we can only recover $K_s$ and $\alpha$ from degrees $\ell>0$.
Once $K_s$ is known, we can estimate $K_d$ from degree $\ell=0$.
Put simply, once we know the contribution of the specular component, we can recover the diffuse component exactly, up to measurement error.
This leads to an important insight:
\textbf{we can reduce uncertainty evaluation to the specular component parameters $K_s$ and $\alpha$.}
This simplification comes naturally in the frequency domain, because we can limit our analysis to degrees $\ell > 0$, but would be challenging in the angular domain.

\subsubsection{Power spectrum}
We propose to further simplify our analysis using the power spectrum of the spherical harmonics coefficients of the incoming and outgoing radiance.
The power spectrum of the incoming radiance $L$ can be computed per degree from the spherical harmonic coefficients
\begin{equation}
    S_{L}(\ell) = \sum_{m=-\ell}^\ell L_{\ell m}^2
\end{equation}
The power spectrum is invariant to rotations of the coordinate system, which, in our context, makes it invariant to slight perturbations of the normals at each point. Using \autoref{eq:specular}, we can compute the power spectrum for a hypothetical outgoing radiance for degrees $\ell>0$, given a set of parameters $\psi$ as
\begin{align}
    S_{B^\psi}(\ell) &= \sum_{m=-\ell}^\ell (K_s e^{-(\alpha \ell)^2}L_{\ell m})^2 \\
                  &= K_s^2 e^{-2(\alpha \ell)^2}\sum_{m=-\ell}^\ell L_{\ell m}^2
                  = K_s^2 e^{-2(\alpha \ell)^2} S_{L}(\ell)
    \label{eq:powerspectrum_brdf}
\end{align}
We can now express the error function $d$, with $\psi = (K_s,\alpha)$, based on the error between the observed power spectrum and predicted power spectrum
\begin{equation}
    \boxed{
    d(B, r(\psi, L)) = \sum_{\ell=1}^{\ell^{*}} (S_{B}(\ell) - K_s^2 e^{-2(\alpha \ell)^2} S_{L}(\ell))^2}
    \label{eq:lightweight_objective}
\end{equation}
$\ell=0$ is left out, because it does not contribute to uncertainty. Using the power spectrum reduces the number of summands in this equation from $(\ell^*+1)^{2} - 1$ to $\ell^{*}$, because we do not need to evaluate \autoref{eq:specular} for all coefficients. This lets us parallelize the evaluation of the objective for many parameter combinations and locations on the surface, making evaluation near instantaneous (see Section~\ref{sec:experiments}).

The combination of entropy with frequency analysis allows us to derive similar observations to \citet{Ramamoorthi_Hanrahan_2001}, as summarized in the Supplement, Section D.1. For example, uncertainty is high when the lighting contains only low frequencies, because many roughness values would give similar results (\autoref{fig:entropy}, center). Low albedo also causes high uncertainty, because less light is reflected, amplifying the relative effect of measurement noise (\autoref{fig:entropy}, right).

\subsection{Fitting Spherical Harmonic Coefficients}
\label{sec:method_fitting_spherical_harmonics} 

To analyze the reflection function in the frequency domain, we need a spherical harmonics transform for the incoming and outgoing radiance samples. However, the samples of outgoing radiance are captured in a multi-view capture setup, meaning they are sparse and irregularly distributed on the upper hemisphere. The typical approach would be to fit spherical harmonic coefficients using least-squares (see Supplement). However, we find that naive least-squares fitting is insufficient to recover BRDF parameters in the frequency domain (detailed in the Supplement). We address this by adding a weighted $L2$ norm on the spherical harmonic coefficients in the least-squares formulation
\begin{equation}
(\mathbf{Y}^\intercal\mathbf{Y} + \lambda \mathbf{W})\mathbf{c} = \mathbf{Y}^\intercal \mathbf{f},
\label{eq:sh_lsq_regularized}
\end{equation}
where $\mathbf{f}$ is a vector of $n$ discrete samples of the signal, $\mathbf{Y}$ is a matrix of size $n \times (\ell^{*} + 1)^2$ containing the spherical harmonics evaluated at the sampling points, and $\mathbf{c}$ is a vector of the $(\ell^{*} + 1)^2$ coefficients we want to find. $\mathbf{W}$ is a diagonal weight matrix. If constant, this weight matrix leads to poor fitting as the regularization is too strong on the lower spherical-harmonics degrees. We set the weight equal to $e^\ell$, increasing the strength of regularization for higher spherical harmonics degrees. This is supported by the observation that many natural images have a power spectrum with exponential decay~\cite{fleming2003illumination}. Intuitively, this regularizer encourages filling unknown regions with low-frequency information.
We weigh the samples based on their elevation angle $\theta$ with $|\cos(\theta)|$, since observations at grazing angles are more likely to present measurement error.

\paragraph{Aliasing and ringing} The incoming signal should be bandlimited to roughly $v^{\frac{1}{2}}$ to avoid aliasing, where $v$ is the number of input views.  We ensure this is the case for the incoming radiance by filtering the input environment map with the BRDF function for $\alpha'=v^{-\frac{1}{2}}$. Our uncertainty and parameter estimation is therefore approximate for $\alpha < \alpha'$. A more in-depth analysis is provided in the Supplement. In practice, we find that the loss in accuracy is acceptable in comparison to the gains in efficiency (\autoref{sec:quality_entropy}). Ringing, which is often an issue for spherical-harmonics based approaches~\cite{sloan2017deringing}, does not impact our method, because we do not represent the BRDF directly as spherical harmonics, but fit a parametric BRDF model to the first $l^*$ degrees in the power spectrum. The pre-filtering and regularization of spherical-harmonics fitting ensure that ringing does not impact how we handle incoming and outgoing radiance.

%% file: sections/4_experiments.tex
\section{Experiments}
\label{sec:experiments}

\paragraph{Implementation}
We implement our method in PyTorch. Mitsuba 3 \cite{jakob2022mitsuba3} is used to render results and as a reference method. Each timing result is computed on an NVIDIA RTX4090 GPU and 32-core Intel Core i9-14900KF CPU.
Code for our experiments is available at \url{https://github.com/rubenwiersma/svbrdf_uncertainty}.

\paragraph{Datasets and tasks}
The experiments are performed on Stanford ORB~\cite{kuang2023stanfordorb} and on synthetic scenes. Stanford ORB is an in-the-wild benchmark that contains fourteen objects, each captured three times in different scenes (lighting environments), selected from a total of seven scenes. The lighting environment is captured through a chrome ball and stored as a lat-long environment map. Each object is also scanned in a separate stage, providing high-quality geometry. For our evaluation, we focus on the (SV)BRDF recovery step and use the provided geometry.

For the synthetic benchmark, we gathered a selection of fifteen objects with spatially varying BRDF textures for base color, roughness and metallicity. The benchmark is rendered with four environment maps representing varying challenges: two indoor scenes, one outdoor scene with a clear sky and sun, and one overcast outdoor scene.
Because ground-truth material textures are available for the synthetic scenes, we can quantitatively evaluate our optimization results and validate uncertainty directly on the optimized parameters. The objects are rendered in Mitsuba at $512\times 512$ resolution and $256$ samples per pixel. Renders and ground-truth views for both Stanford ORB and the synthetic benchmark are included in the Supplement.

\subsection{Quality of Inverse Rendering}
As a first validation, we are interested in the performance of the spherical-harmonics model (SH model) evaluated in the angular domain, including our proposed additions for shadowing and masking, on the task of inverse rendering. This validates that the SH model is a relevant proxy for more recent and exact models in inverse rendering and lets us evaluate the model without the approximations required for the power spectrum simplification. We will also use the SH model in our applications of entropy, where this model acts as a baseline to compare the applications with.
We optimize PBR material textures to match renderings from multiple views (512x512 textures of base color, roughness, metallicity), UV-mapped to the surface. We use a simple mapping between the PBR materials and the parameters used in the SH model, which is described in the Supplement.
The parameters are optimized using gradient descent with an Adam optimizer that minimizes an L1 reconstruction loss and a TV-norm regularizer in texture space weighted by \num{1e-2}.

\input{tables/comparisons/acquisition_stanford_orb}
\input{tables/comparisons/synthetic_benchmark}

\input{figures/stanford_orb_inverse_rendering}
\input{figures/synthetic_inverse_rendering}

Our main comparison target is Mitsuba 3 \cite{jakob2022mitsuba3}, a complete differentiable path tracer. When optimizing with Mitsuba, we sample one ray per pixel in each view and randomly select a batch of rays per optimization step for better convergence. Each primary ray is sampled $32$ times to  sample reflection directions well.
The relighting results on Stanford ORB are shown in \autoref{tab:benchmark_stanfordorb} and visual comparisons in \autoref{fig:stanford_orb_inverse_rendering}. We find that our SH model performs on par with Mitsuba but requires only $1/10$ of the time despite little optimization on our side, as we use pure PyTorch. Our power spectrum variant, which ignores Fresnel, shadowing, and masking, still performs on par with NVDiffRec, with a 30-fold speedup compared to Mitsuba. 
On the synthetic benchmark, we compare the optimized textures to the ground-truth textures used to render the dataset. The results are presented in \autoref{tab:benchmark_acquisition_synthetic} and visual comparisons in \autoref{fig:synthetic_inverse_rendering} and the Supplement. The SH Model achieves $1.36$dB lower PSNR than Mitsuba on average. The results for Mitsuba are the best-case scenario, given that the dataset was rendered with Mitsuba, using the same BRDF model used for material estimation. Moreover, the dataset contains particularly challenging objects for our method (e.g., with significant interreflections). Split out over base color (Mitsuba: $20.54$dB, SH: $18.77$dB), roughness (Mitsuba: $17.84$dB, SH: $14.71$dB) and metallicity (Mitsuba: $12.94$dB, SH: $13.78$dB), we find that roughness is particularly challenging for the SH model. This is expected, because the effect of roughness is approximated by the SH model. Given that our goal is to estimate uncertainty, we find the performance of the SH-model on this stress-test acceptable. In \autoref{sec:sharing}, we discuss how to improve these results using information sharing (`global sharing' rows in \autoref{tab:benchmark_acquisition_synthetic}).

\input{figures/stanford_orb_entropy}
\input{figures/synthetic_entropy}
\subsection{Quality of Approximate Entropy}
\label{sec:quality_entropy}
Our method simplifies the reflection equation, so it can be evaluated in the power spectrum of spherical harmonics.
This yields significant performance gains, but involves approximations. To study the validity of these approximations, we compare the entropy computed with our method to the entropy computed using Mitsuba and a variant of our method that evaluates the reconstruction error in the angular instead of the frequency domain. This allows us to evaluate the signal-processing framework for inverse rendering without additional approximations.
While using a path-tracing renderer like Mitsuba to compute entropy is impractical for most use cases, it provides a close proxy for `ground-truth' uncertainty.

We compute entropy for every scene in Stanford ORB with each approach and use the Pearson correlation coefficient $\rho$ between the entropy maps as a quality measure. We do not compare exact entropy values, as both methods compute the loss in a different space, making it difficult to exactly compare the resulting probability distributions. For Mitsuba, we discretize the parameter space into $8\times 8\times 8$ parameter combinations (roughness, metallicity, base color) and $8 \times 8$ for our method (specularity and roughness, see explanation below \autoref{eq:specular}). We find a high correlation between entropy computed with Mitsuba and both our method evaluated in the angular domain $\rho=0.94$ and in the power spectrum $\rho=0.88$ as shown in Table~\ref{tab:benchmark_uncertainty}. This validates that the entropy we compute with our power spectrum based model is representative of one computed with a physically-based renderer, which confirms that our approximations are permissible for uncertainty.
Visual comparisons of entropy in \autoref{fig:stanford_orb_entropy} confirm these findings. Further, as our power spectrum approximation is parallelizable, we obtain extreme speedups over both the angular-domain approach ($3,320\times$) and Mitsuba ($700,000\times$) making uncertainty estimation practical.
For a higher (and more accurate) sampling density of $16 \times 16 \times 16$, Mitsuba takes roughly $47$ minutes, the angular-domain approach $13.2s$, and our power-spectrum method only $0.0006s$.

\subsection{Entropy and Error}
\label{sec:experiments_uncertainty_synthetic_dataset}
An important part of our motivating argument is that high certainty is predictive of low error. This makes uncertainty useful for various applications, e.g., information sharing across areas, or inpainting of uncertain areas using a diffusion model. We argue that regions with high uncertainty are more likely to converge to an incorrect parameter, simply because an incorrect parameter yields a similar error to the correct parameter. However, an uncertain region does not necessarily converge to an incorrect result: the correct parameter choice could still have the maximum likelihood, even by a small margin, or the optimization could converge on the correct choice by chance. We therefore do not expect a perfect correlation between error and entropy. At most, one could expect a correlation of $0.5$.

We study the relationship between entropy computed with our power-spectrum model and error in the texture maps estimated with Mitsuba. We intentionally test the relationship between different models to show the generalizability of our approach.
In the last row of \autoref{tab:benchmark_acquisition_synthetic} we show the correlation scores, which are $0.22$ on average and $0.27$ for the photostudio scene.
We also show qualitative results in Figure~\ref{fig:synthetic_entropy}. These results are optimized on images rendered under the ``rural asphalt road'' environment. We observe a match between high entropy- and high error regions. To illustrate the usefulness of entropy, we show a mask of all regions with above-average error and a mask of regions with above-average entropy. Such masks could be used to remove erroneous regions. We show that the entropy-based mask covers nearly all regions of high error, while still being sparse. This demonstrates that entropy can serve as an a-priori guide to reduce error.

\input{tables/comparisons/uncertainty_stanford_orb}

\section{Applications}
\label{sec:applications}
To demonstrate use of uncertainty and practicality of our method, we propose three applications of entropy in the context of SVBRDF recovery: guiding capture, sharing information, and inpainting uncertain regions with a diffusion model.

\subsection{Guiding Capture}
\begin{figure}[t]
    \centering
    \includegraphics[width=\linewidth]{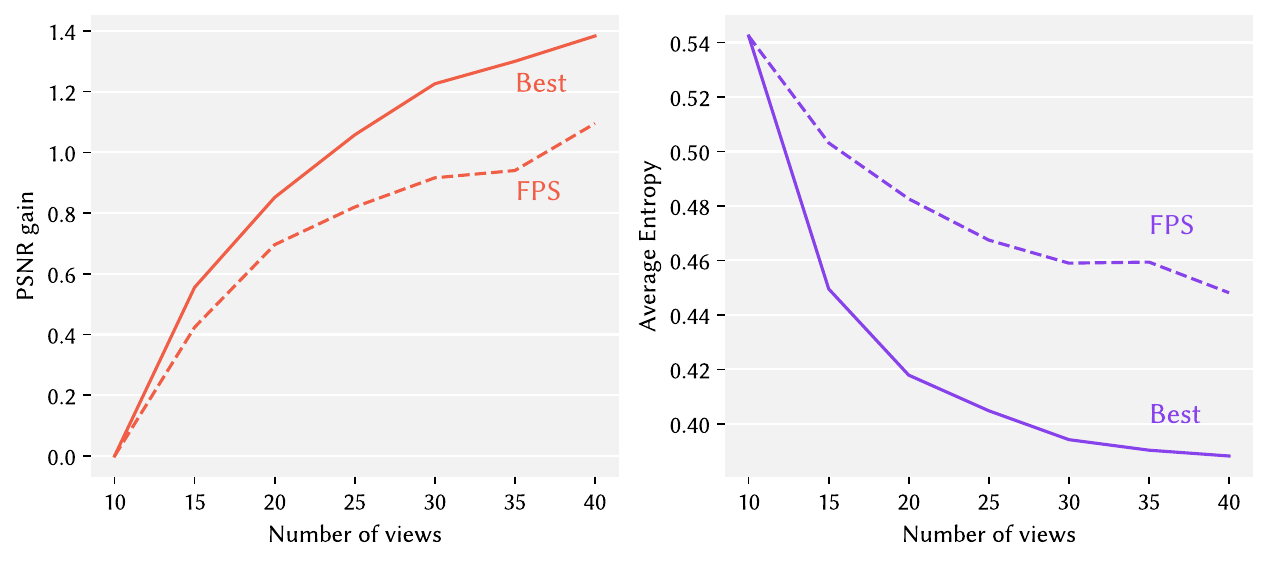}
    \caption{Plots of PSNR gain and entropy for varying input view counts on the synthetic benchmark with the Museum environment. We observe a significant improvement of our best-view selection (Best) over farthest point sampling (FPS) on a paired t-test ($p=$\num{1e-8}). Using farthest point sampling requires 60\% more views compared to using our entropy-based capture guidance.}
    \label{fig:application_capture}
    \Description[Two line graphs side-by-side, showing the PSNR gain and Average Entropy over the number of views. A solid line is plotted for the method labeled Best and a dashed line for FPS.]{The figure shows two line graphs side-by-side. The plots show the PSNR gain and Average Entropy over the number of views. A solid line is plotted for the method labeled Best and a dashed line for FPS. In the PSNR gain plot, both Best and FPS are increasing and there graph for Best is above the FPS graph. In the Average Entropy plot, both graphs are decreasing and the graph for Best is below the FPS graph.}
\end{figure}

Entropy can be applied during capture as a measure to define the most informative viewpoints. This is challenging, because we do not know what information can be captured from a viewpoint before capturing it. Therefore, we propose an approach to compute the \textit{expected} entropy (i.e., information gain) from adding a new viewpoint: we compute the entropy for hypothetical observations $B'_\psi$ for a parameter combination $\psi$, given incoming radiance observed from a hypothetical set of viewpoints, $L'$, sampled from the known lighting. The entropy for a hypothetical observation is weighed by the probability of the parameters responsible for that observation, $\bar{p}(\psi | B, L')$ from \autoref{eq:normalizedlikelihood}. The result is expected entropy
\begin{equation}
    E[H] = \sum_{\psi \in \Psi} \bar{p}(\psi | B, L') H(B'_\psi, L').
\end{equation}
We use our power spectrum approximation to accelerate this computation: we first estimate the spherical harmonics coefficients for $L'$ and compute the hypothetical radiance $B'$ only in the power spectrum, based on $S_{L'}$. 
A set of viewpoints is then selected with a greedy strategy, which we call \textit{best-view selection}. Starting with $v'$ evenly spaced viewpoints, we compute the expected entropy for $V$ candidate viewpoints and select the viewpoint with the lowest expected entropy. This is repeated until $v$ desired viewpoints are selected. We compare this algorithm with farthest point sampling (FPS) of the $V$ candidate viewpoints as a baseline. Our hypothesis is that the best-view selection yields better performance, given the same number of viewpoints. In our experiment, we use $V=100$ candidate points and $v'=10$ starting points. The same starting points are used for the best-view selection and FPS baseline. We evaluate both approaches on the synthetic dataset and compare the PSNR gain and the average entropy over all shapes using the additional views (see \autoref{fig:application_capture}). We observe a considerable improvement in both the PSNR gain and entropy: the number of views required to reach +1 PSNR gain with the FPS baseline is 60\% higher than the number of views the best-view selection requires. We also compare the results with a paired t-test (pairing the same scenes) and see a statistically significant improvement between the best-view selection and the FPS baseline ($p=$\num{1e-8}).

This application shows the importance of our efficient power spectrum approximation, as it involves a nested entropy computation. Using our method, capture guidance requires roughly one second \emph{per added viewpoint}, while this would take roughly 45 minutes in the angular domain and more than nine days using Mitsuba.

\begin{figure}[t]
    \centering
    \includegraphics[width=0.9\linewidth]{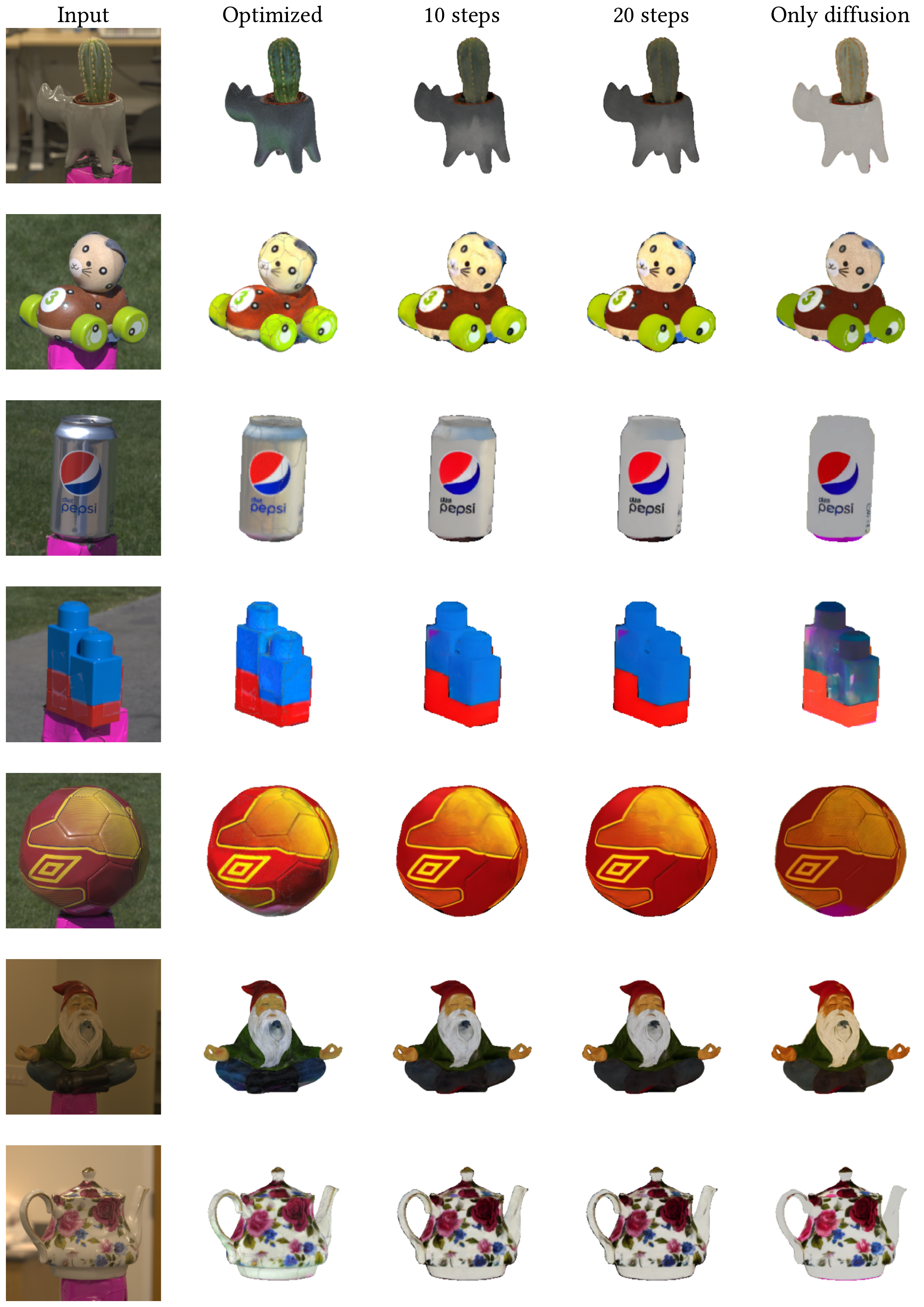}
    \caption{Inpainting low-certainty results with the RGB$\leftrightarrow$x diffusion model~\cite{zheng2024rgbx}. We show the result of the optimization, results for inpainting and a full diffusion model prediction on base color. Results are shown with 10/20 steps of complete diffusion after 40/30 steps of inpainting, striking a balance between preserving the information from the optimization and using the diffusion model to inpaint uncertain areas.}
    \label{fig:application_diffusion}
    \Description[A grid of figures showing the base color inpainting results using a diffusion model on Stanford ORB.]{The figure shows a grid of results for the base color inpainting using a diffusion model. There are seven objects (ceramic plant pot, toy car, soda can, plastic building blocks, soccer ball, gnome statue, and teapot) and for each object, the input image is shown alongside the optimized result, diffusion inpainting with 10 steps and 20 steps and only diffusion. The diffusion results are off in color, while the 10- and 20-steps results match the color of the optimized column, while being more even.}
\end{figure}

\subsection{Sharing Information}
\label{sec:sharing}
Uncertain regions on the surface could be improved with information from more certain regions. Prior works typically use a regularizer,  enforcing smoothness of the parameters over the surface (e.g., by adding a TV-norm term \cite{munkberg2022extracting, hasselgren2022shape, sun2023neural}). Information is consequently shared locally. However, a smoothness term does not distinguish between certain and uncertain points, blurring regions that should actually be preserved. Given our measure of uncertainty, we can develop a more informed strategy: we would like to share information from the most certain points to less certain points with a similar appearance and allow for sharing to happen with non-neighboring points. 
We choose a set of certain points, $C$ (local minima in the uncertainty map $H$). Next, we add a term to the regularizer that enforces other points' parameters to be similar to the selected points:
\begin{equation}
\mathcal{L}_{S} = \frac{1}{T} \sum_{p} \sum_{q \in \mathcal{C}} w_{pq}||\psi_p - \psi_q||^2,
\end{equation}
where $T$ is the number of discrete points on the surface (in our experiments, the number of texels), $\psi_p$ is the parameter at point $p$, and $w_{pq}$ is a weight based on the similarity of the materials at points $p$ and $q$. Since we do not know a priori whether two regions consist of the same material, we use the KL-divergence between the probability distributions $P_p$ and $P_q$ over the possible material parameters as a similarity metric
\begin{equation}
    w_{pq} = \exp\left[ -\frac{D_{KL}(P_p || P_q)^2}{2\tau^2} \right].
\end{equation}
The parameter $\tau$ determines the selectiveness in finding similar regions. We set $\tau$ as a fraction of the average KL-divergence between $C$ and all other points ($0.4$ in our experiments).
The results of this approach are compared with the baseline models using a TV-norm in \autoref{tab:benchmark_acquisition_synthetic}. We observe a +1.5 dB PSNR improvement for the SH model and +0.5 dB for Mitsuba. Visual comparisons reflect these results (\autoref{fig:synthetic_inverse_rendering} and \autoref{fig:stanford_orb_inverse_rendering}, global sharing). We only show results for the SH model as the results for Mitsuba with global sharing are visually similar, albeit slightly better.

\subsection{Diffusion Model Guidance}
We can use the material map estimated with an optimization method, together with our uncertainty map to extract material maps with a diffusion model~\cite{ho2020denoising}. As a proof of concept, we inpaint regions in the estimated albedo map with high uncertainty using blended diffusion~\cite{Avrahami_2022_CVPR}. For the diffusion model, we use RGB$\leftrightarrow$x~\cite{zheng2024rgbx}, which estimates material maps in image space for shaded images. Following the blended diffusion process, we perform $50$ diffusion steps and replace the diffusion result with the optimized material map for the first few steps, followed by $n$ diffusion steps on the full material map. In \autoref{fig:application_diffusion}, we show results on Stanford ORB and the base color parameter for four different variants: no inpainting, inpainting followed by $10$ and $20$ full diffusion steps, and full diffusion. We observe that the results with $10$ full diffusion steps (i.e., more influence of optimization results) strike a good balance between the correct parameter values and the diffusion model result.

\subsection{Optimization: Initialization}
We show that our fast estimate of BRDF parameters can be used as a high-quality initialization for other inverse rendering approaches.
In this experiment, we initialize an optimization with Mitsuba using the textures from the SH Model. This is optimized for one epoch, using only $16$ samples per pixel (half the samples we used for the other experiments) and the results are included in \autoref{tab:benchmark_stanfordorb} (last row). This single epoch only takes $11$s, resulting in a total of $16$s, combined with the initialization. This is over three times faster than the $52$s required with random initialization and achieves better performance (+$0.3$dB PSNR-H).

%% file: tables/comparisons/acquisition_stanford_orb.tex
\begin{table}[tb]
\centering
\setlength{\tabcolsep}{3pt}
\scriptsize
\caption{Benchmark Comparison: Stanford ORB \cite{kuang2023stanfordorb} Relighting. Each method has access to the ground-truth lighting and geometry.
See supplement for additional methods and different acquisition conditions.}

\label{tab:benchmark_stanfordorb}
\resizebox{1.0\linewidth}{!}{
\begin{tabular}{lccccc}
\toprule
&  PSNR-H$\uparrow$ & PSNR-L$\uparrow$ & SSIM$\uparrow$ & LPIPS$\downarrow$ & Time \\\midrule

NVDiffRec~\cite{munkberg2022extracting} & 24.319 & 31.492 & 0.969 & 0.036 & 142.14s \\

Mitsuba~\cite{jakob2022mitsuba3}  & $\mathbf{26.601}$  & $\mathbf{34.195}$  & $\mathbf{0.977}$ &  $0.032$   & $52.49$s \\
\textbf{SH Model} & $26.525$  & $33.796$  & $\mathbf{0.977}$ &  $\mathbf{0.029}$   & $5.07$s \\

\textbf{SH Model - Power Spectrum}  &      $24.494$      &      $31.215$      &     $0.971$      &       $0.035$       & $\mathbf{1.78}$s \\

\midrule
SH Model init + Mitsuba (1 epoch) & $\mathbf{26.918}$  & $\mathbf{34.386}$  & $\mathbf{0.978}$ &  $0.031$   &     $16.07$s      \\

\bottomrule
\end{tabular}
}
\end{table}

%% file: tables/comparisons/synthetic_benchmark.tex
\begin{table}[tb]
\centering
\setlength{\tabcolsep}{3pt}
\scriptsize
\caption{Synthetic benchmark: Mitsuba vs SH Model - PSNR to ground-truth and correlation between SH-model entropy and error for Mitsuba.}

\label{tab:benchmark_acquisition_synthetic}
\resizebox{1.0\linewidth}{!}{
\begin{tabular}{lcccccc}
\toprule
 & Photo Studio  &  Museum  &  Overcast  &  Rural Road  &  Avg. & Time  \\
\midrule
Mitsuba     & $17.07$                   & $17.69$             & $16.35$            & $17.32$                 & $17.11$ &    $74.54$s    \\
\textbf{Mitsuba global sharing}    & $18.07$                   & $18.38$             & $17.96$            & $17.27$                 & $17.92$ &    $77.81$s    \\
\midrule
\textbf{SH Model}   & $15.23$                   & $15.63$             & $15.60$            & $16.55$                 & $15.75$ & $3.09$s    \\
\textbf{SH Model global sharing} & $17.34$                   & $18.10$             & $18.11$            & $18.95$                 & $18.13$  & $3.15$s \\
\midrule
Error-Entropy correlation &   $0.27$               &               $0.20$               &               $0.19$               &               $0.24$               &               $0.22$             &  \\
\bottomrule
\end{tabular}
}
\end{table}

%% file: figures/stanford_orb_inverse_rendering.tex
\begin{figure}
    \centering
    \begin{subfigure}[b]{\linewidth}
        \centering
        \includegraphics[width=\textwidth]{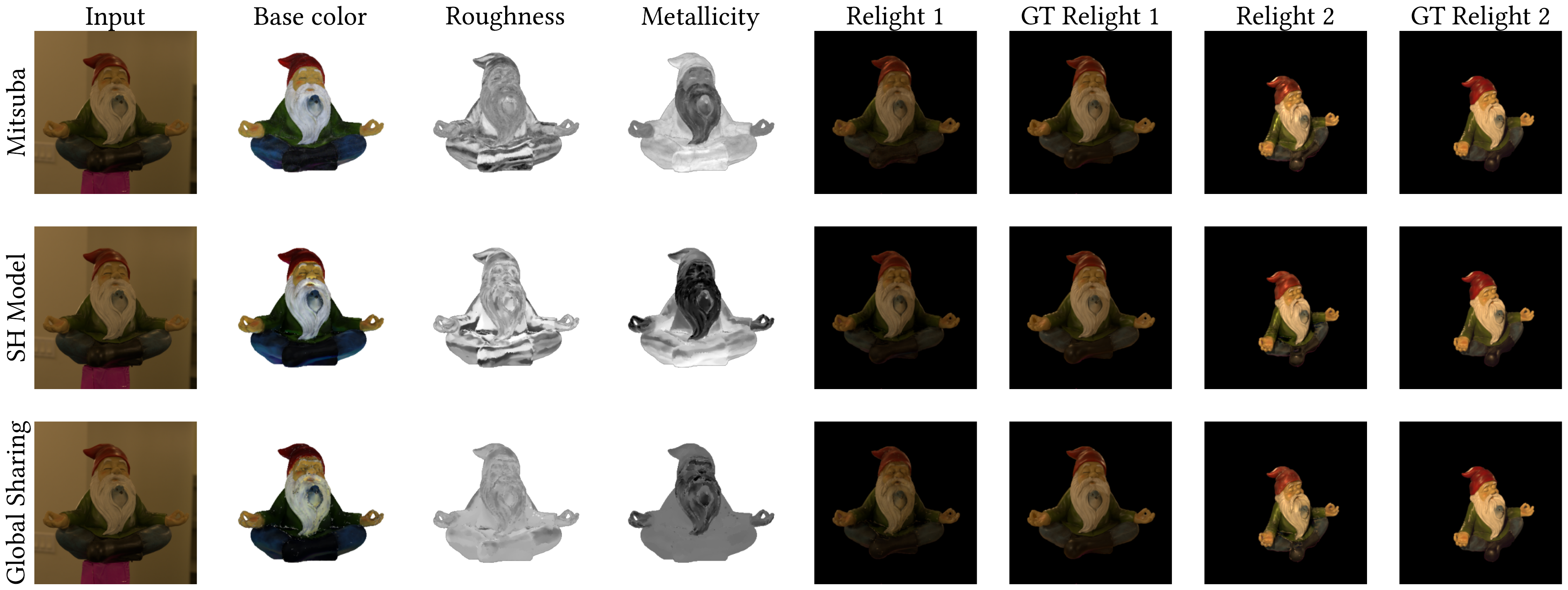}
    \end{subfigure}
    \begin{subfigure}[b]{\linewidth}
        \centering  
        \includegraphics[width=\textwidth]{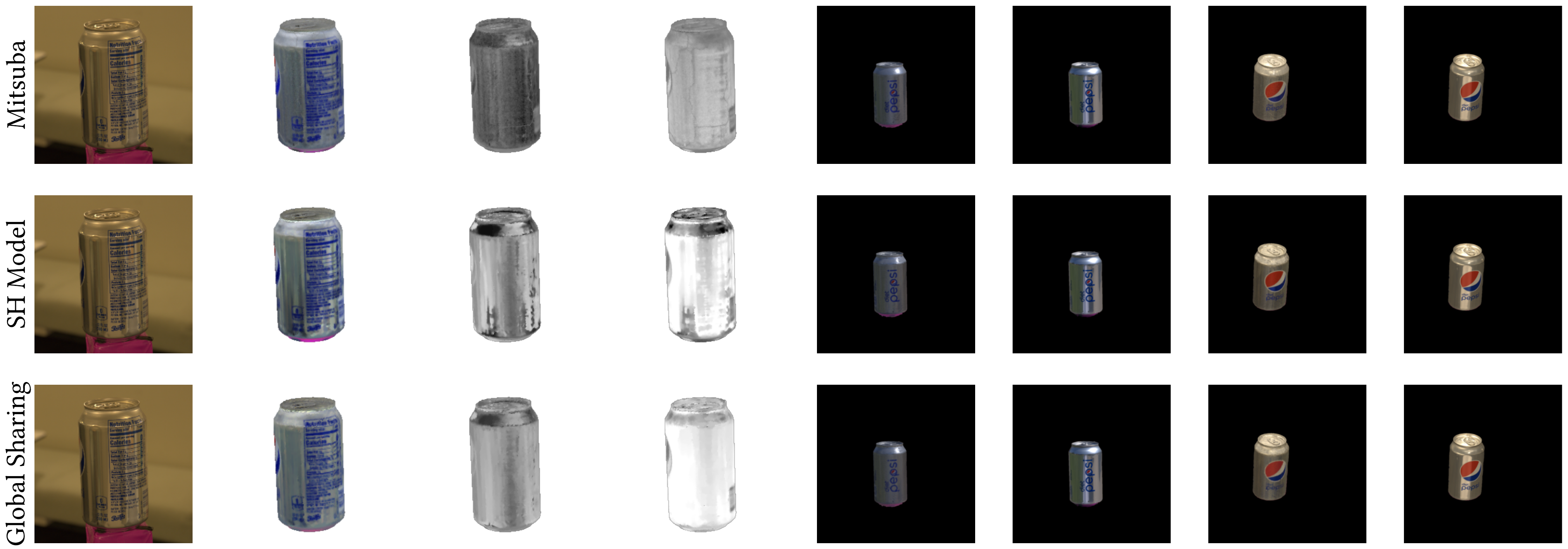}
    \end{subfigure}
    \begin{subfigure}[b]{\linewidth}
        \centering
        \includegraphics[width=\textwidth]{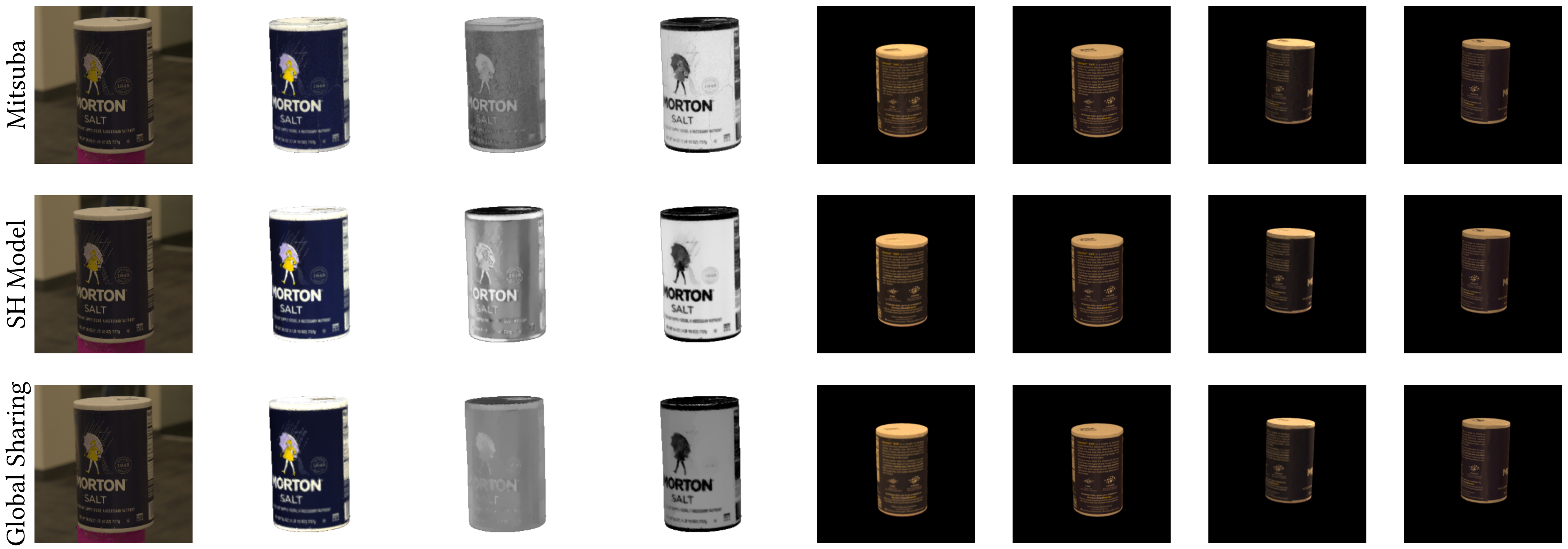}
    \end{subfigure}
    \begin{subfigure}[b]{\linewidth}
        \centering
        \includegraphics[width=\textwidth]{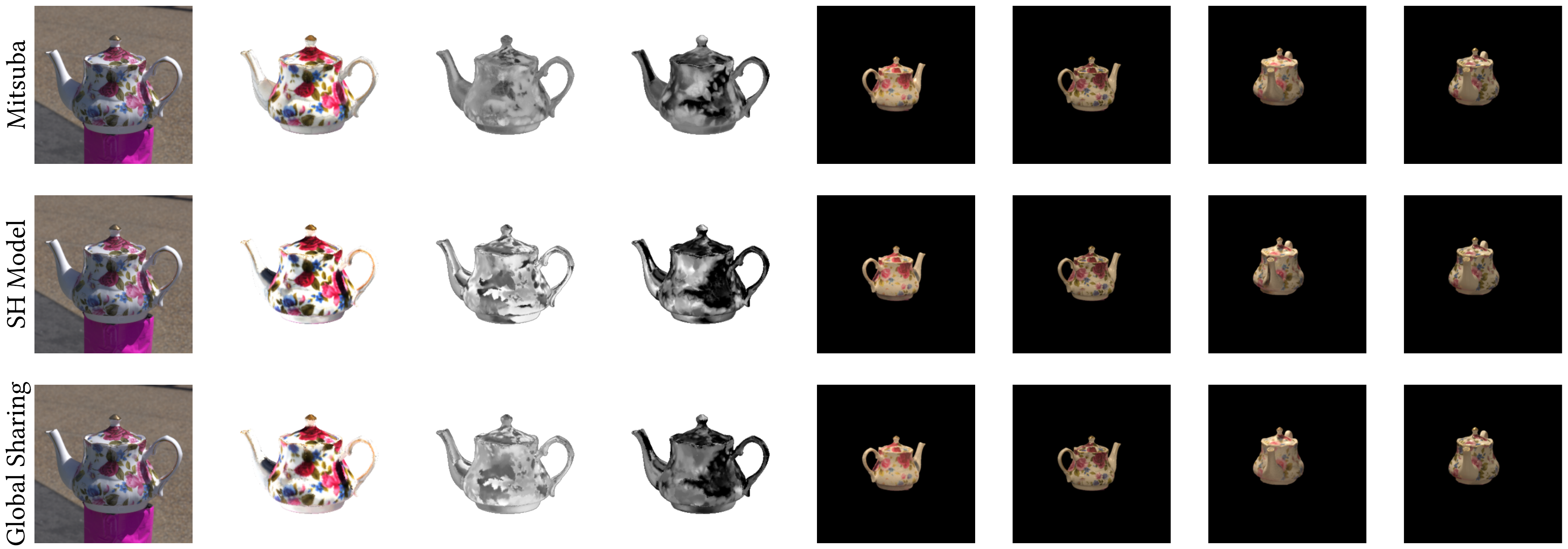}
    \end{subfigure}
    \caption{Material recovery results on Stanford ORB for Mitsuba ($52.49$s on average), the SH Model and our proposed global sharing application ($5.07$s on average).}
    \label{fig:stanford_orb_inverse_rendering}
    \Description[The figure shows a grid of result images on the Stanford ORB benchmark.]{The figure shows a grid of result images on the Stanford ORB benchmark. The columns of the grid are labeled: input, base color, roughness, metallicity, relight 1, ground-truth relight 1, relight 2, ground-truth relight 2. The rows are organized in groups of three, where the groups correspond to objects (gnome statue, soda can, salt container, teapot) and the rows within the group to methods (Mitsuba, SH Model, SH Model with Global Sharing).}
\end{figure}

%% file: figures/synthetic_inverse_rendering.tex
\begin{figure}
    \centering
    \begin{subfigure}[b]{0.49\linewidth}
        \centering
        \includegraphics[width=\textwidth]{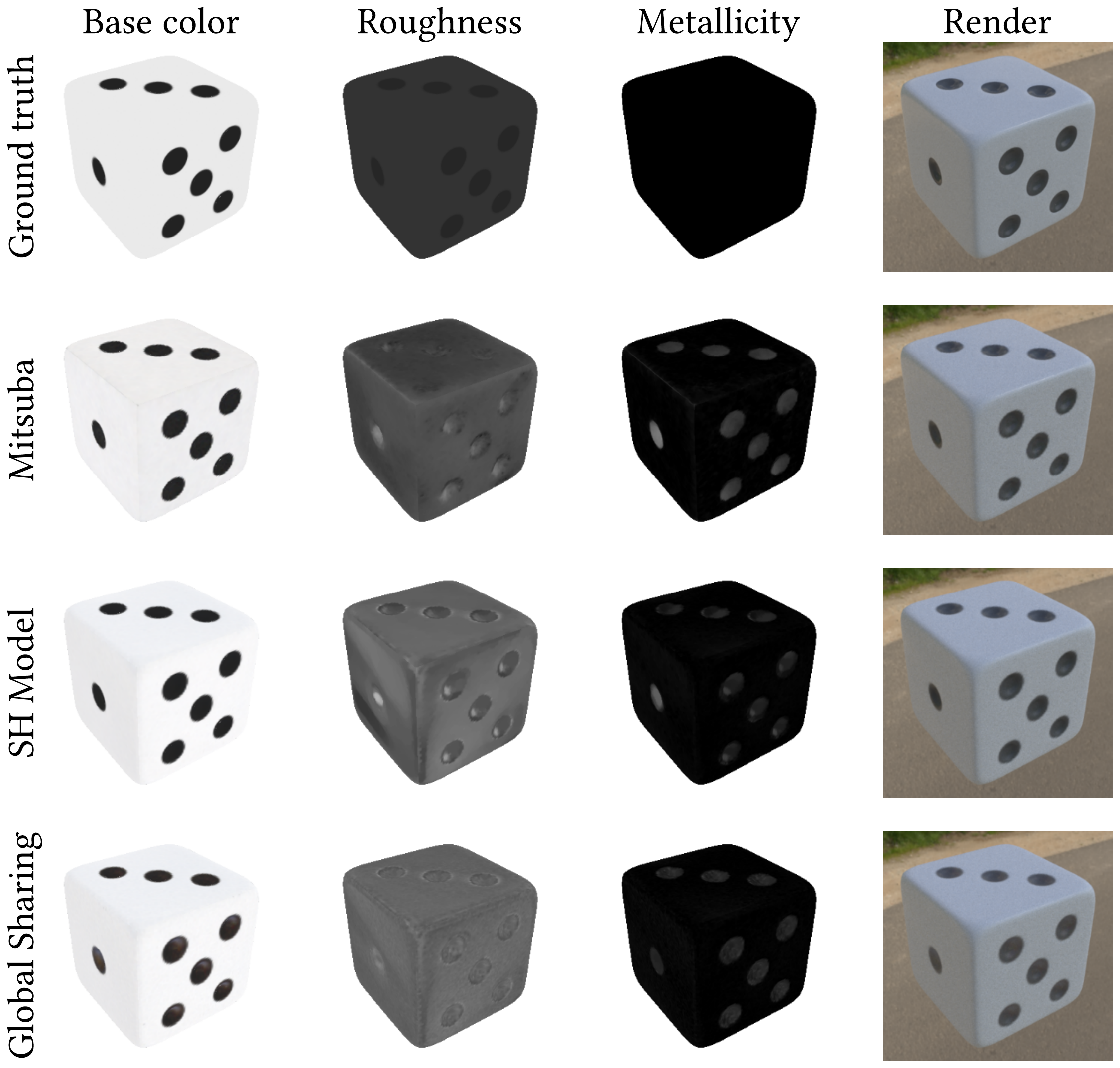}
    \end{subfigure}
    \begin{subfigure}[b]{0.49\linewidth}
        \centering
        \includegraphics[width=\textwidth]{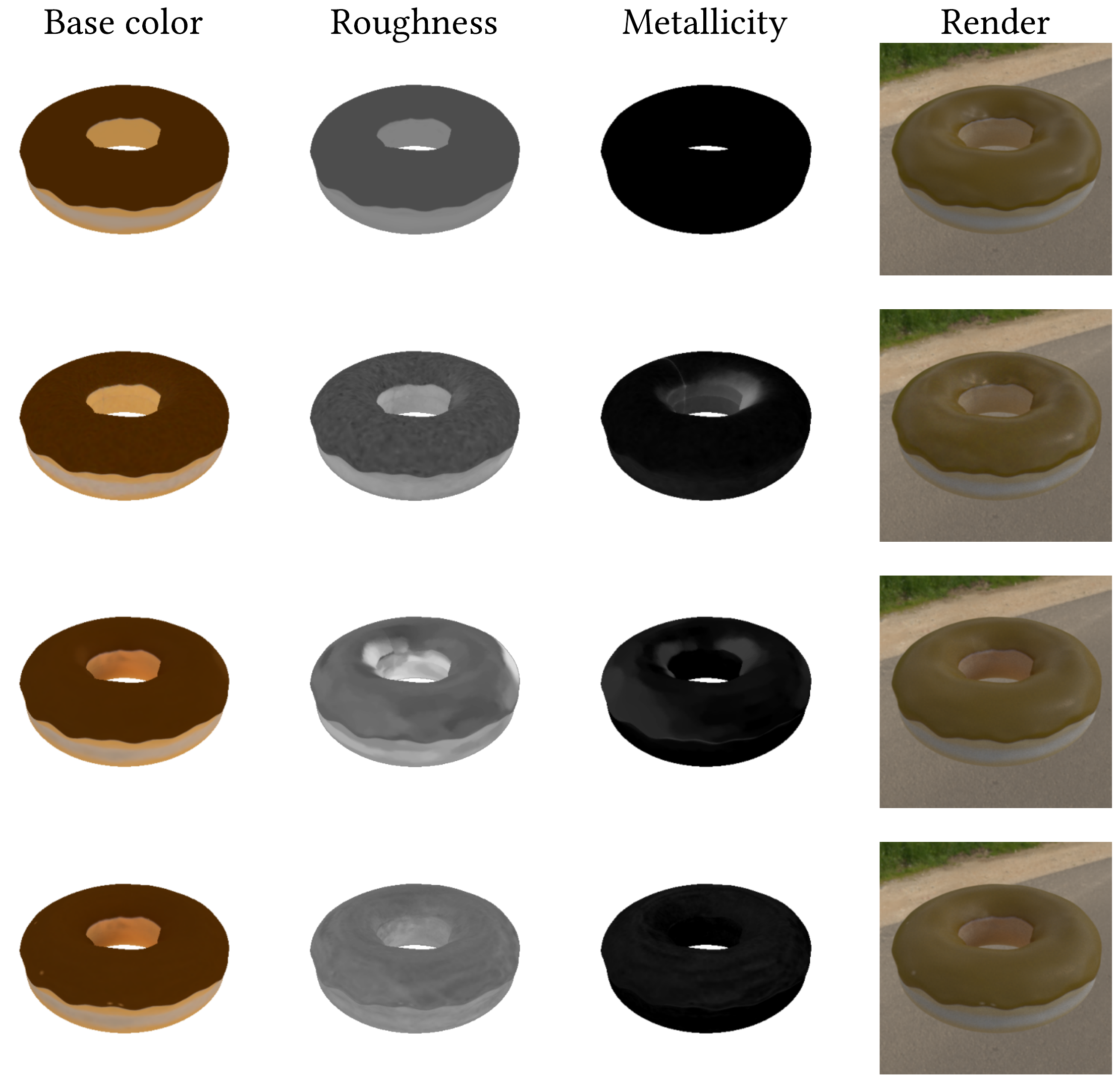}
    \end{subfigure}
    \begin{subfigure}[b]{0.49\linewidth}
        \centering
        \includegraphics[width=\textwidth]{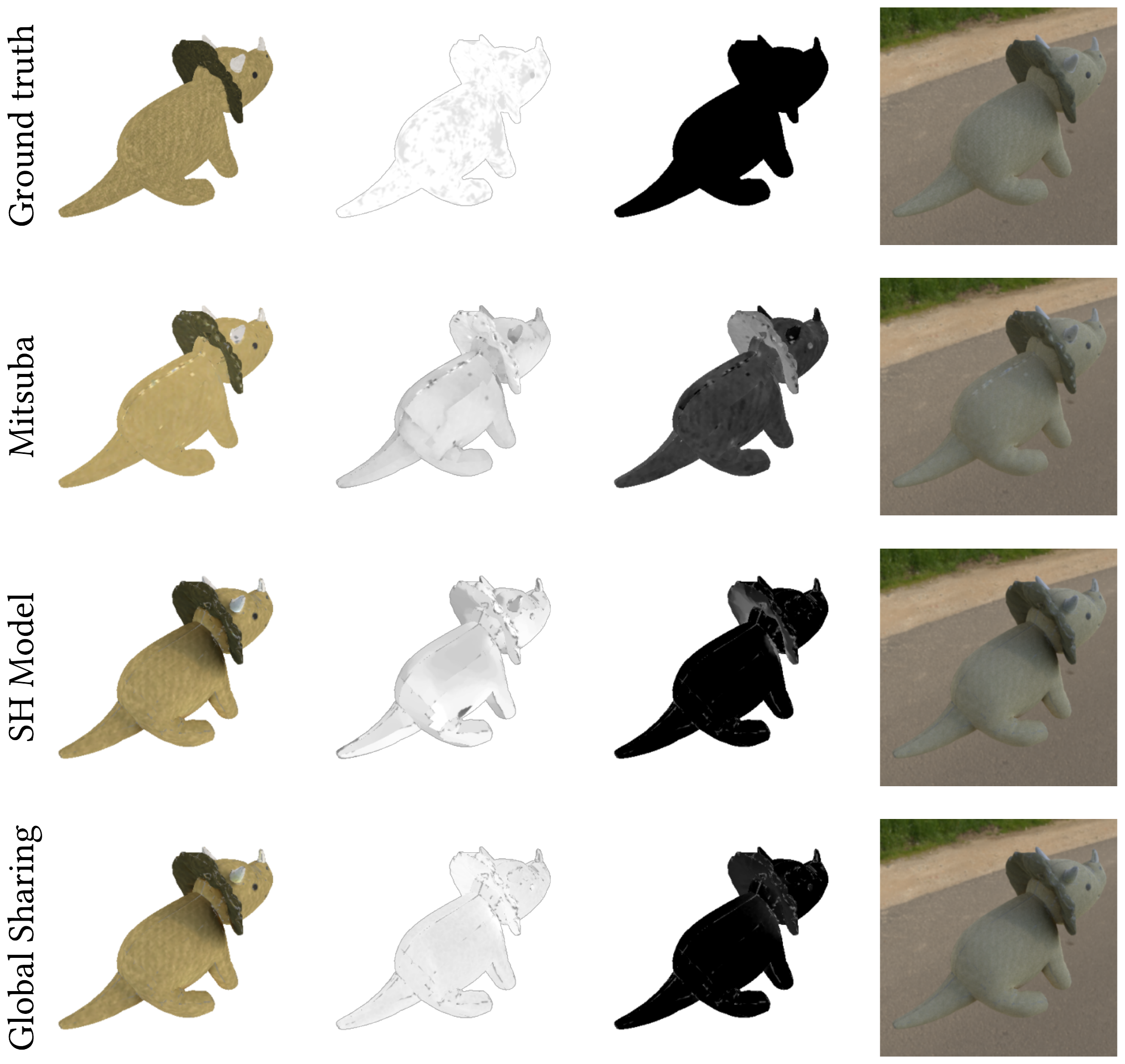}
    \end{subfigure}
    \begin{subfigure}[b]{0.49\linewidth}
        \centering
        \includegraphics[width=\textwidth]{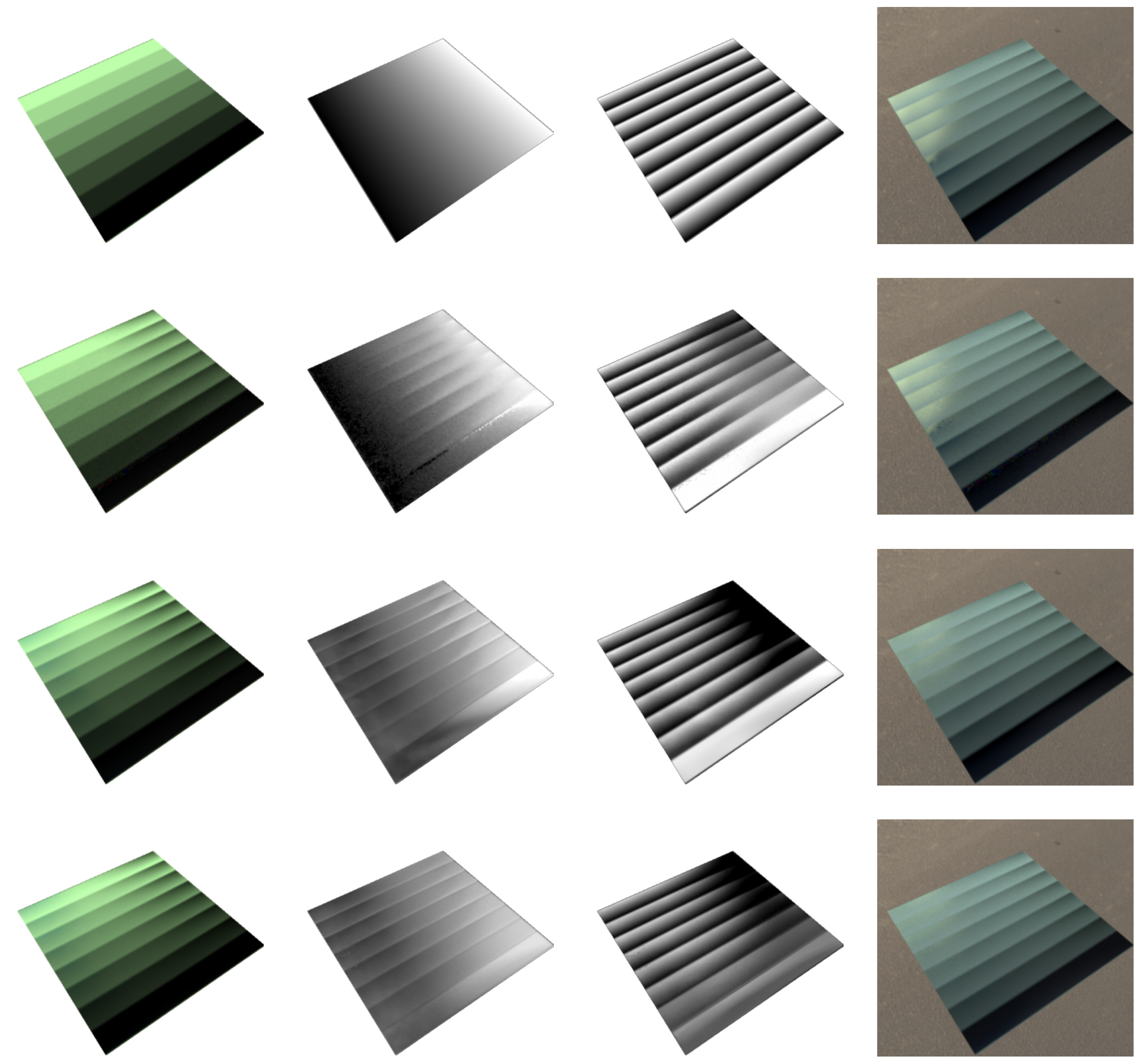}
    \end{subfigure}
    \begin{subfigure}[b]{0.49\linewidth}
        \centering
        \includegraphics[width=\textwidth]{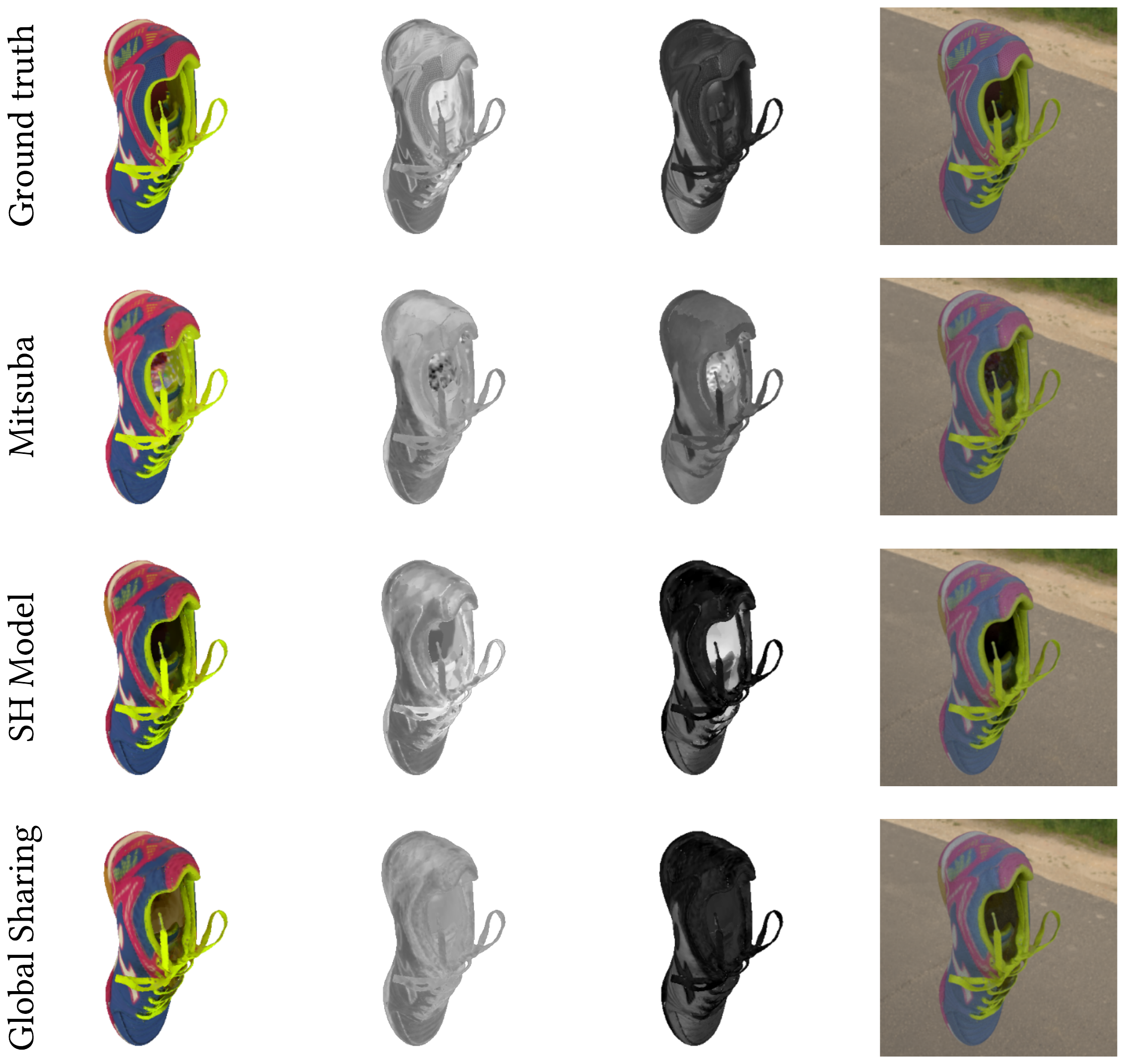}
    \end{subfigure}
    \begin{subfigure}[b]{0.49\linewidth}
        \centering
        \includegraphics[width=\textwidth]{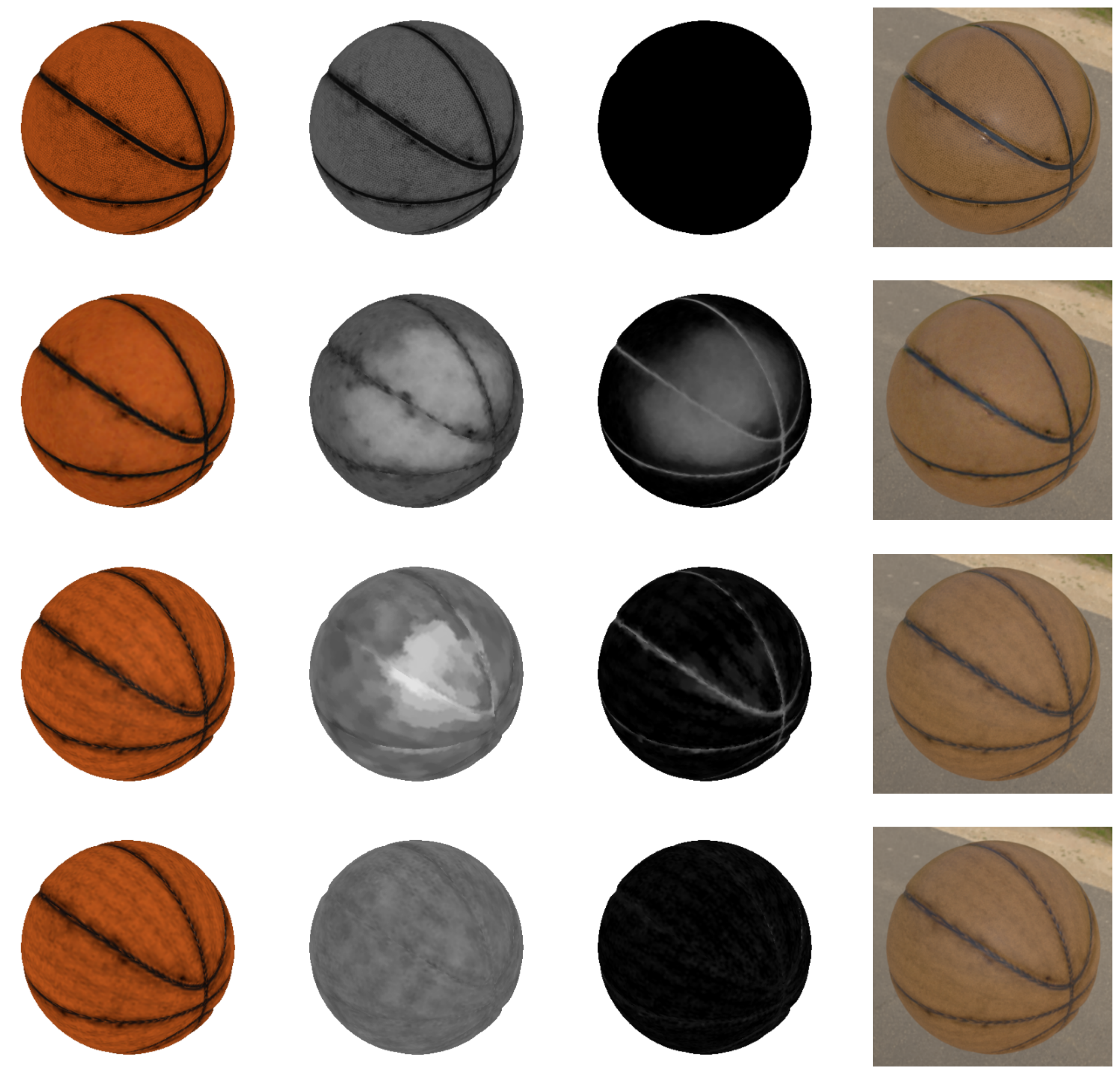}
    \end{subfigure}
    \caption{Material recovery results on the synthetic dataset for Mitsuba, the SH Model, and the SH Model with global sharing enabled by entropy. We see that global sharing helps smooth and improve the results obtained with the SH Model without blurring the maps.}
    \label{fig:synthetic_inverse_rendering}
    \Description[The figure shows a grid of result images on the synthetic benchmark.]{The figure shows a grid of result images on the synthetic benchmark. The grid shows results for six objects (dice, donut, dino plushie, plane, shoe, basketball), organized in a larger two-by-three grid. Each sub-grid shows the resulting base color, roughness, metallicity and a render for the ground truth, Mitsuba, SH Model and SH Model + Global Sharing.}
\end{figure}

%% file: figures/stanford_orb_entropy.tex
\begin{figure}
    \centering
    \begin{subfigure}[b]{0.48\linewidth}
        \centering
        \includegraphics[width=\textwidth]{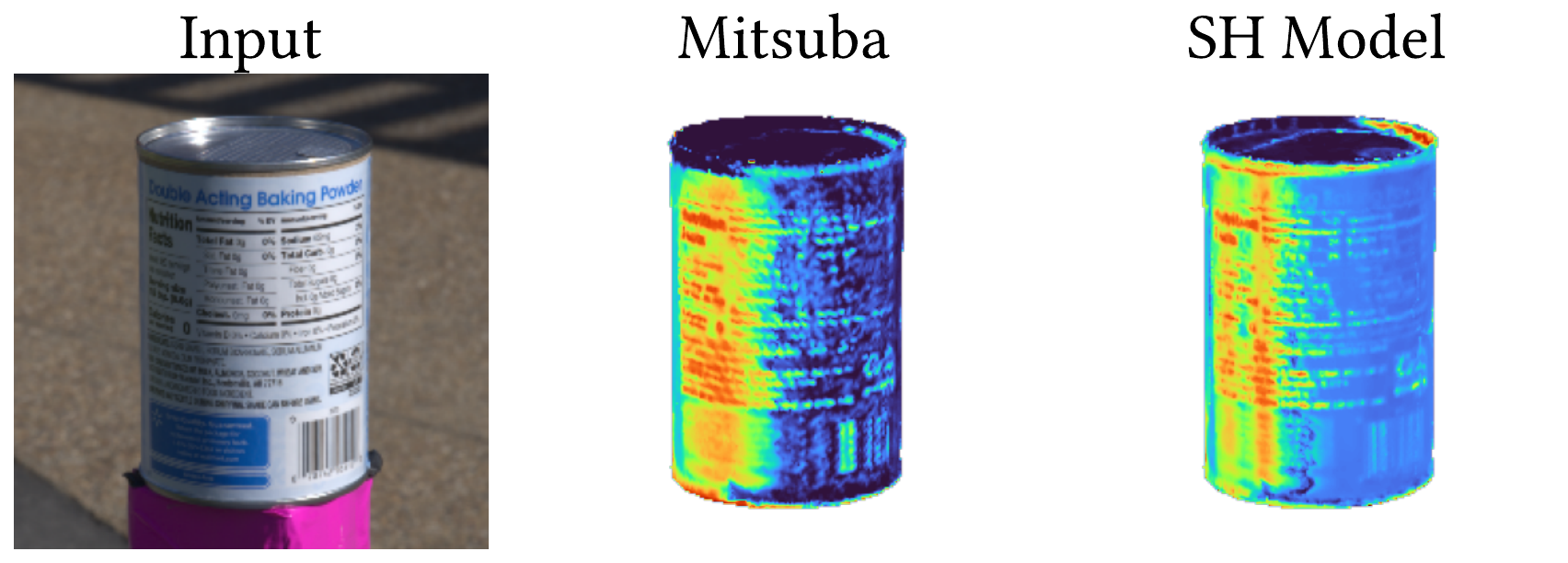}
    \end{subfigure}
    \begin{subfigure}[b]{0.48\linewidth}
        \centering
        \includegraphics[width=\textwidth]{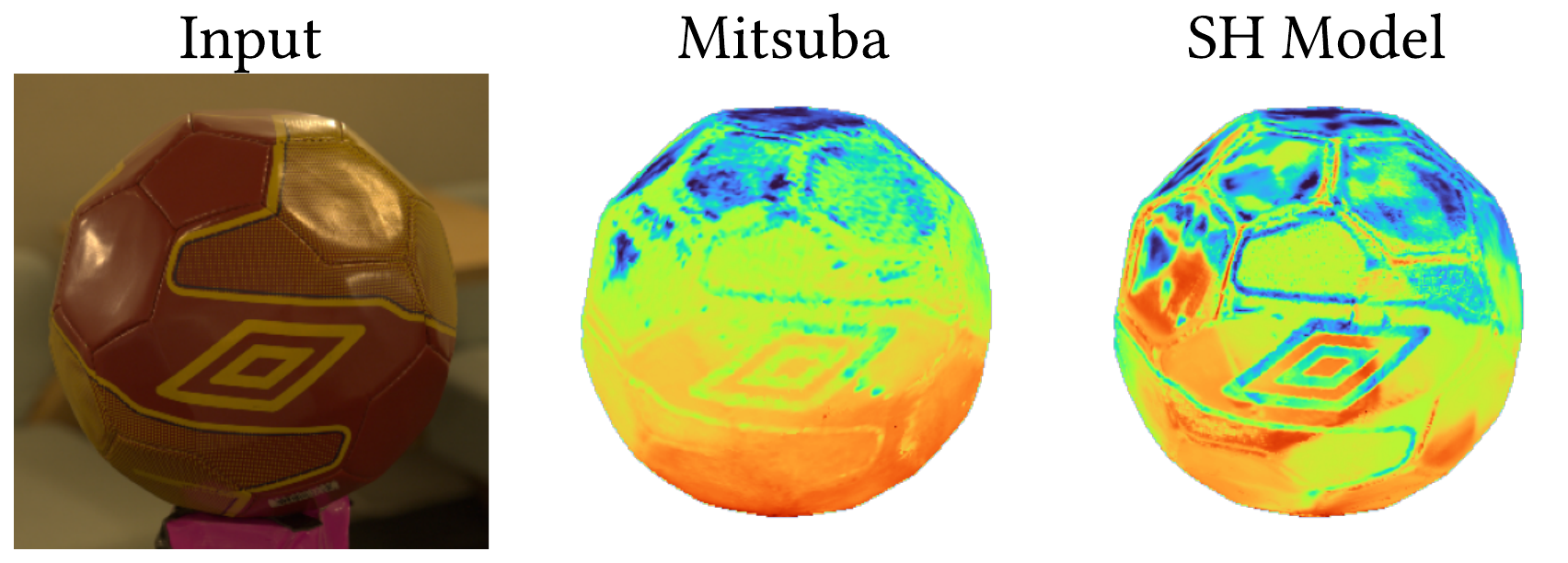}
    \end{subfigure}

    \begin{subfigure}[b]{0.48\linewidth}
        \centering
        \includegraphics[width=\textwidth]{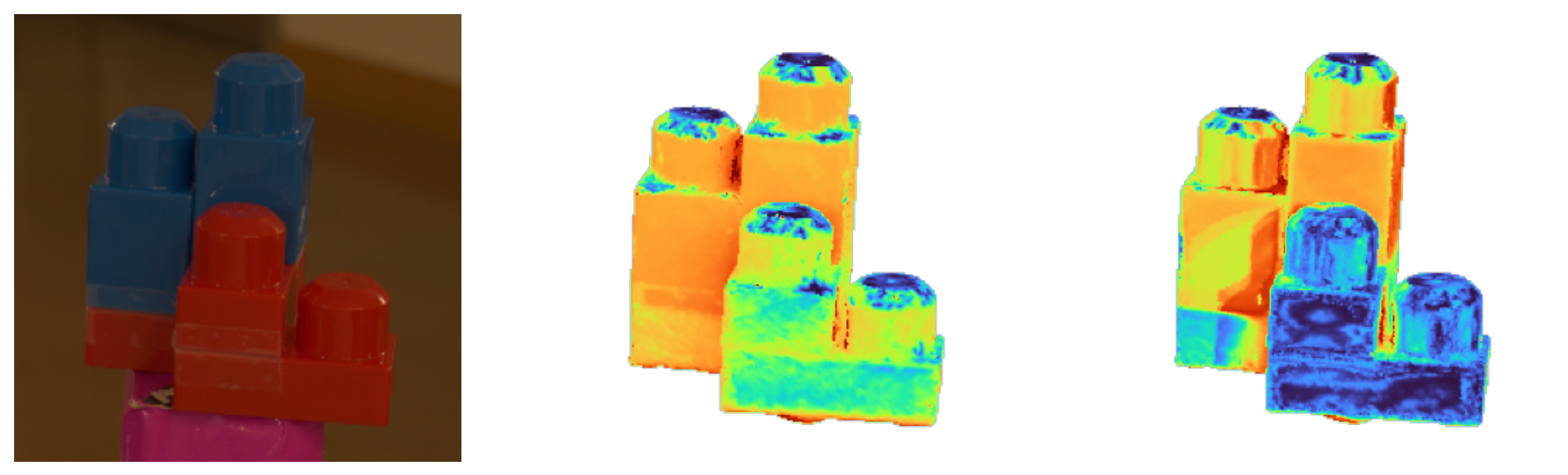}
    \end{subfigure}
    \begin{subfigure}[b]{0.48\linewidth}
        \centering
        \includegraphics[width=\textwidth]{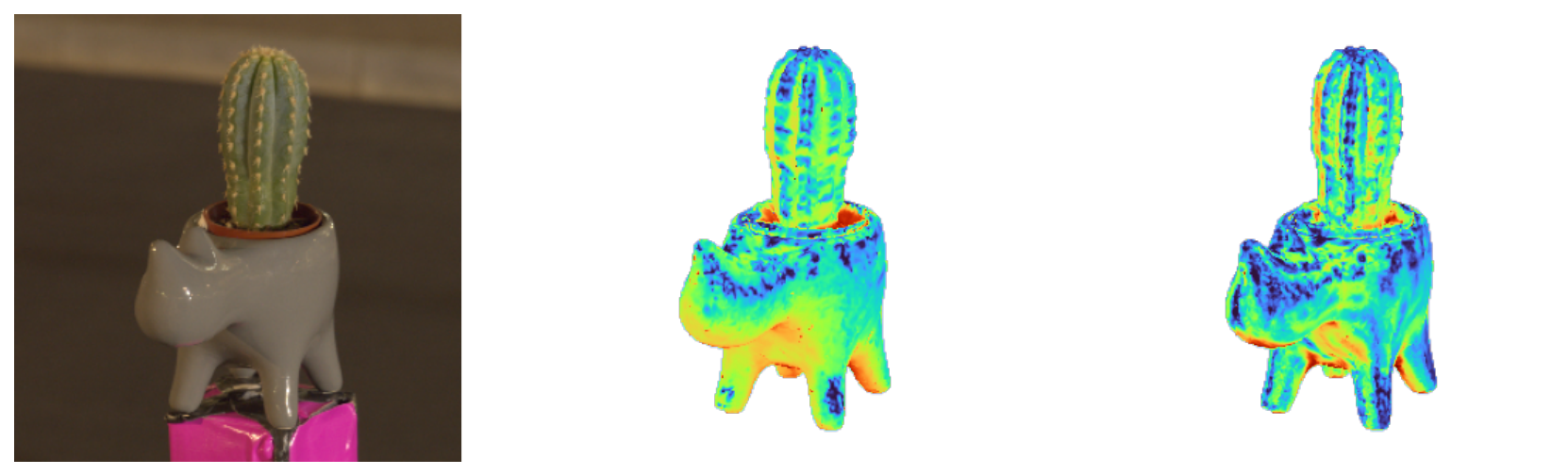}
    \end{subfigure}

    \begin{subfigure}[b]{0.48\linewidth}
        \centering
        \includegraphics[width=\textwidth]{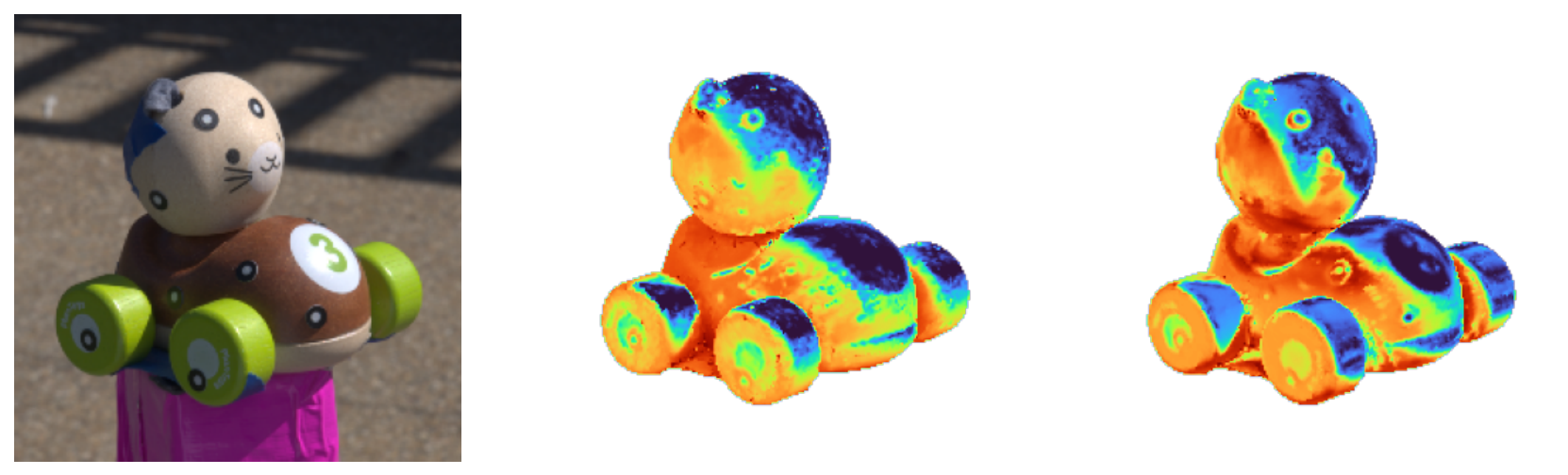}
    \end{subfigure}
    \begin{subfigure}[b]{0.48\linewidth}
        \centering
        \includegraphics[width=\textwidth]{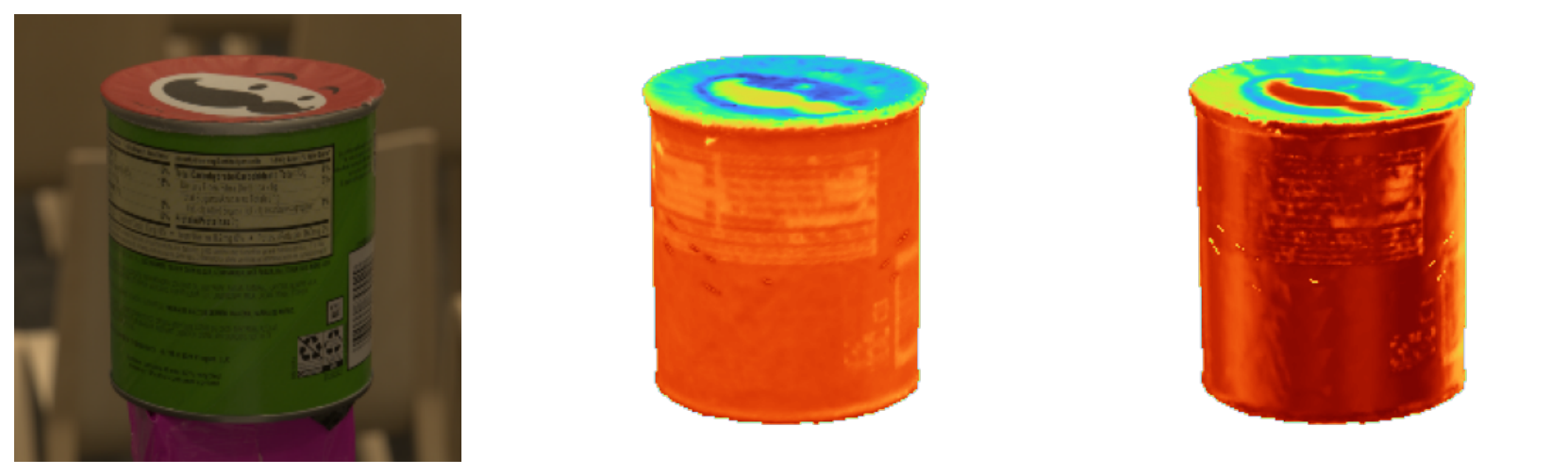}
    \end{subfigure}

    \begin{subfigure}[b]{0.48\linewidth}
        \centering
        \includegraphics[width=\textwidth]{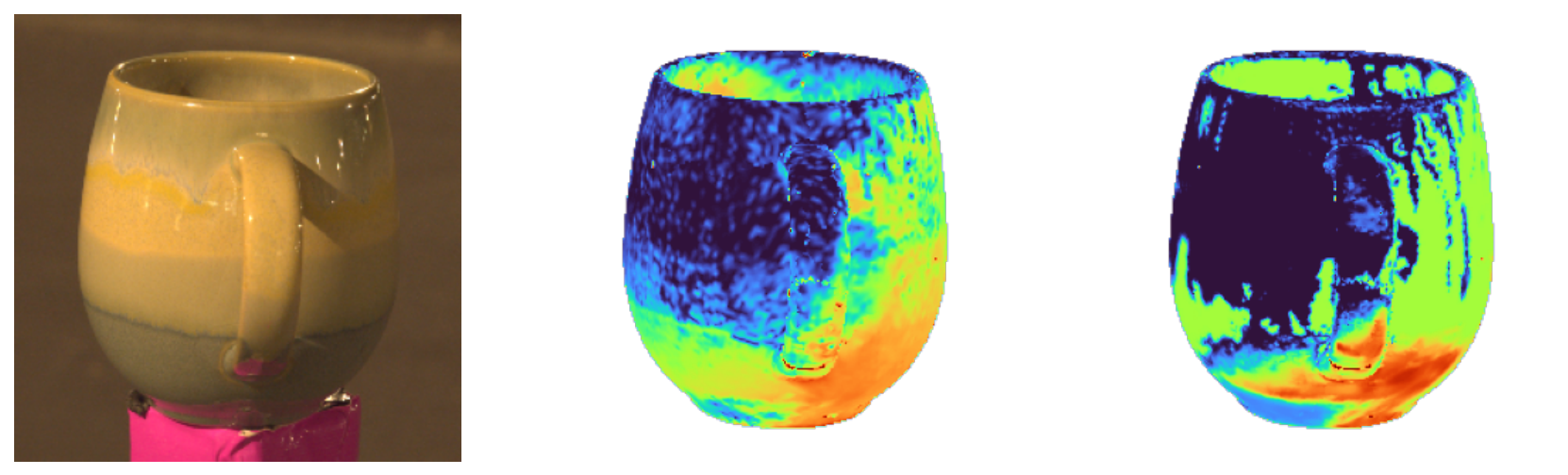}
    \end{subfigure}
    \begin{subfigure}[b]{0.48\linewidth}
        \centering
        \includegraphics[width=\textwidth]{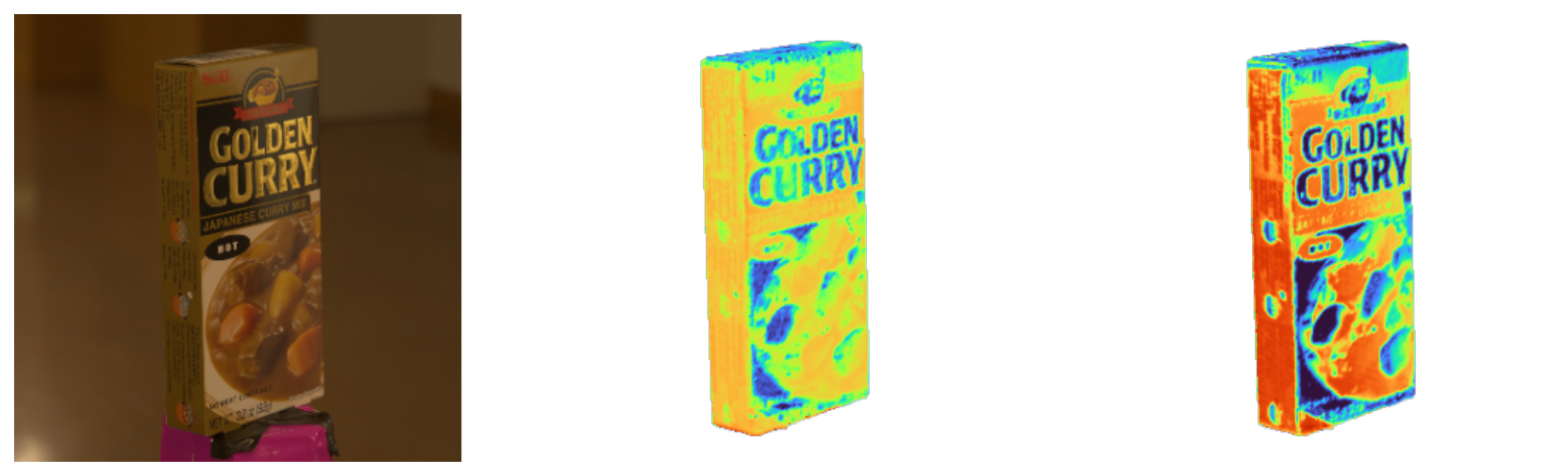}
    \end{subfigure}

    \begin{subfigure}[b]{0.48\linewidth}
        \centering
        \includegraphics[width=\textwidth]{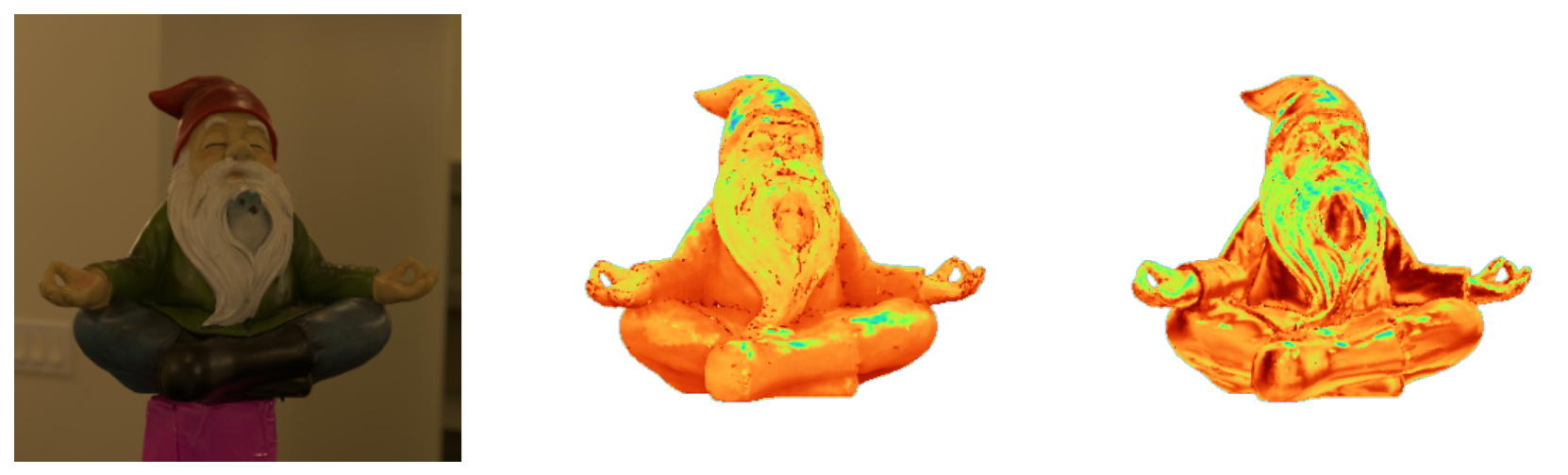}
    \end{subfigure}
    \begin{subfigure}[b]{0.48\linewidth}
        \centering
        \includegraphics[width=\textwidth]{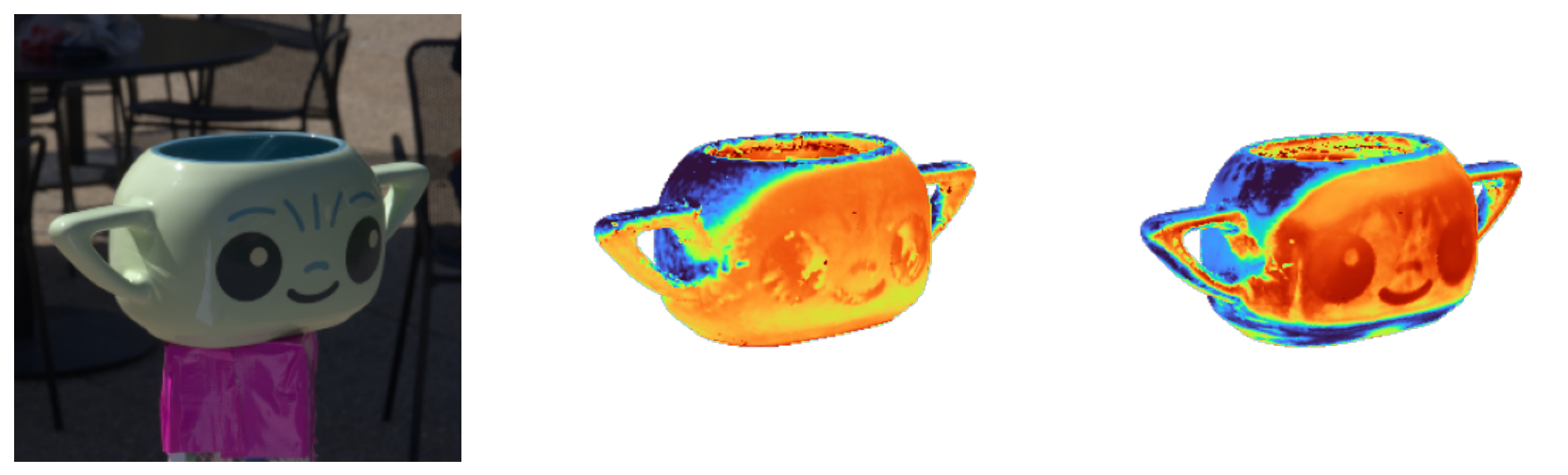}
    \end{subfigure}

    \begin{subfigure}[b]{0.48\linewidth}
        \centering
        \includegraphics[width=\textwidth]{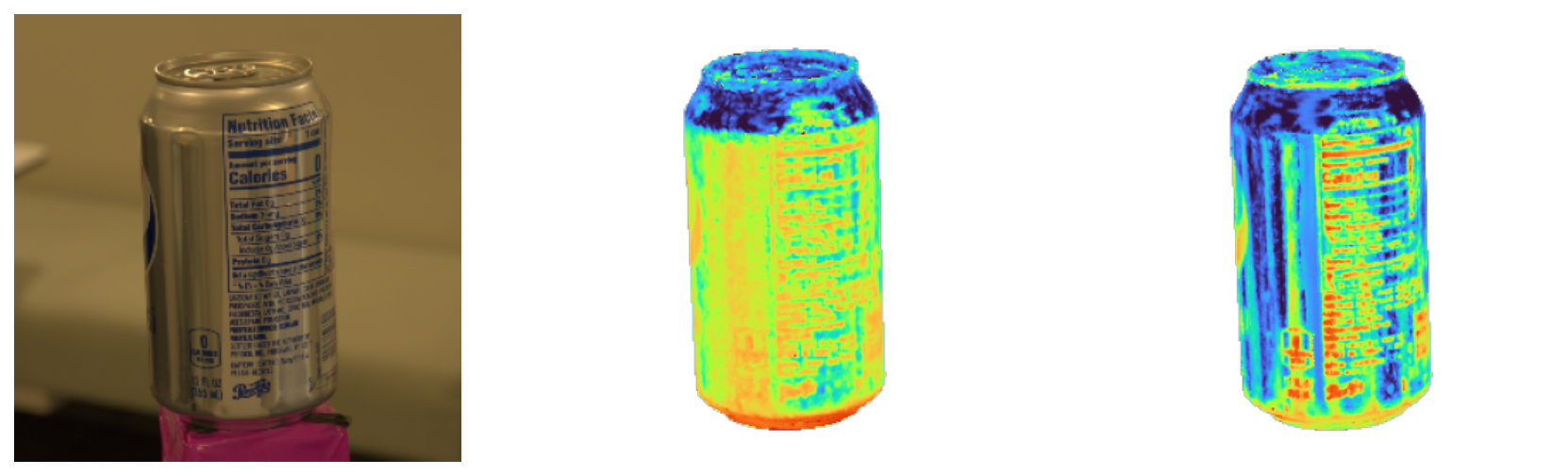}
    \end{subfigure}
    \begin{subfigure}[b]{0.48\linewidth}
        \centering
        \includegraphics[width=\textwidth]{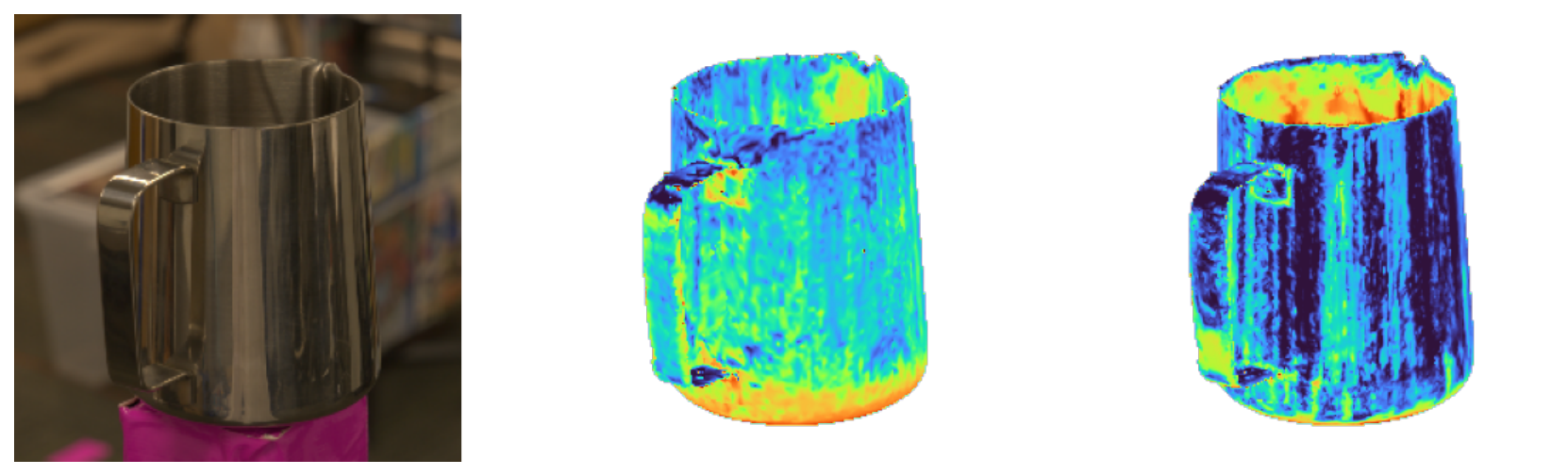}
    \end{subfigure}

    \begin{subfigure}[b]{0.48\linewidth}
        \centering
        \includegraphics[width=\textwidth]{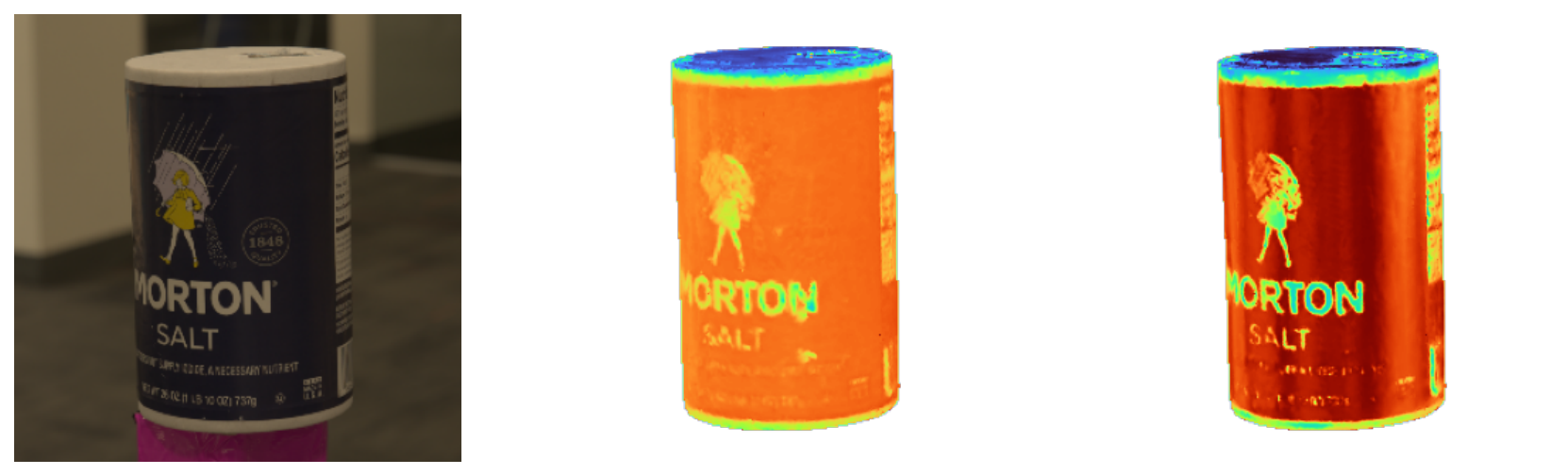}
    \end{subfigure}
    \begin{subfigure}[b]{0.48\linewidth}
        \centering
        \includegraphics[width=\textwidth]{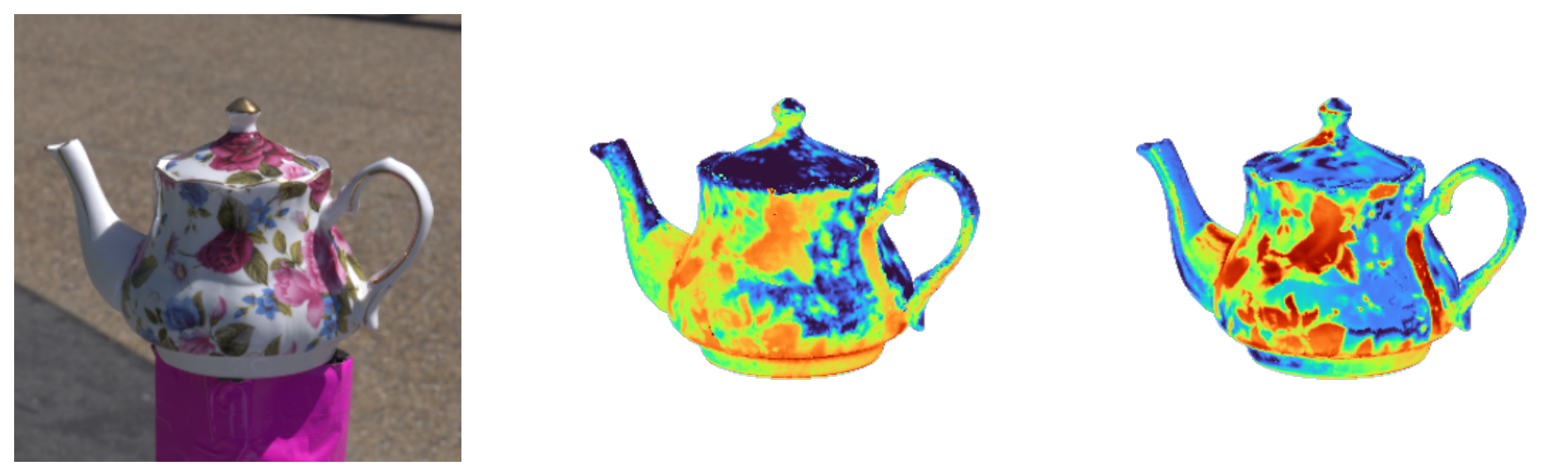}
    \end{subfigure}
    \caption{Entropy estimation results on Stanford ORB for Mitsuba ($5$m$50$s on average) and our method ($0.0005$s on average). This ensures our approximation still produces entropy close to that computed, slower, with Mitsuba }
    \label{fig:stanford_orb_entropy}
    \Description[The figure shows a grid of entropy comparisons on Stanford ORB with Mitsuba and the SH Model.]{The figure shows a grid of entropy comparisons on Stanford ORB with Mitsuba and the SH Model. For each of the fourteen objects, one of the input images is shown, alongside the entropy map for Mitsuba and the SH Model from the same viewpoint. The entropy is represented with a colormap that matches between the Mitsuba and SH Model columns.}
\end{figure}

%% file: figures/synthetic_entropy.tex
\begin{figure*}[tbh]
    \centering
    \begin{subfigure}[b]{0.48\linewidth}
        \centering
        \includegraphics[width=\textwidth]{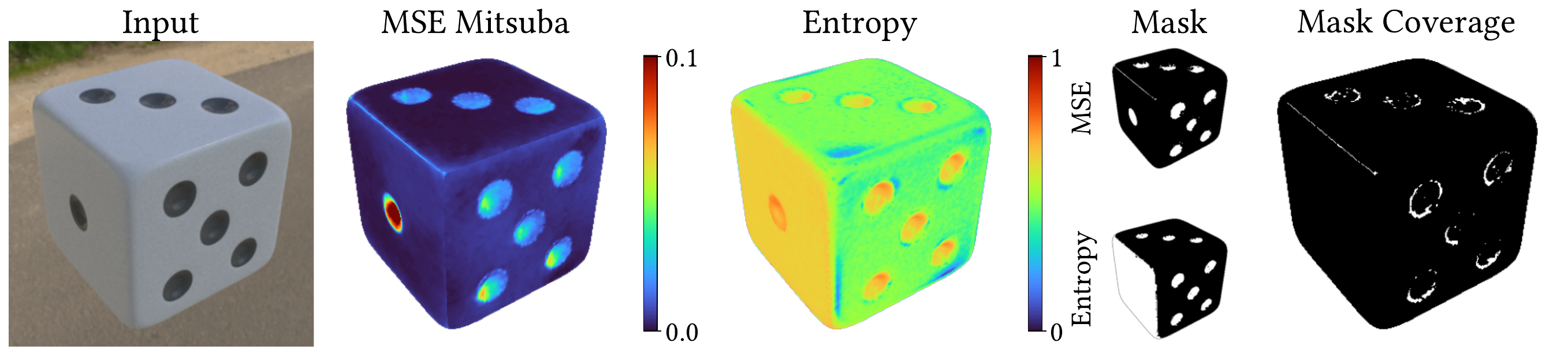}
    \end{subfigure}
    \hspace{0.02\linewidth}
    \begin{subfigure}[b]{0.48\linewidth}
        \centering
        \includegraphics[width=\textwidth]{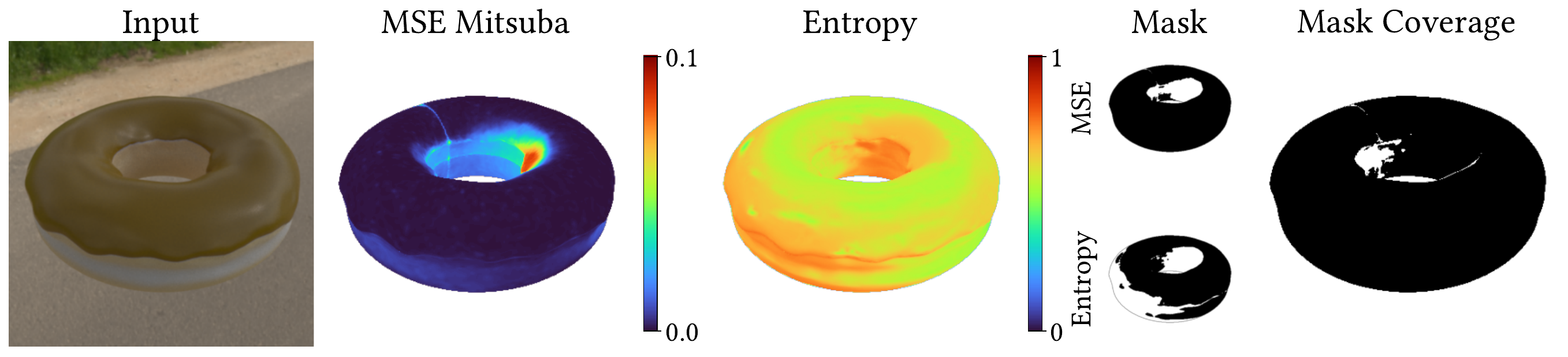}
    \end{subfigure}

    \begin{subfigure}[b]{0.48\linewidth}
        \centering
        \includegraphics[width=\textwidth]{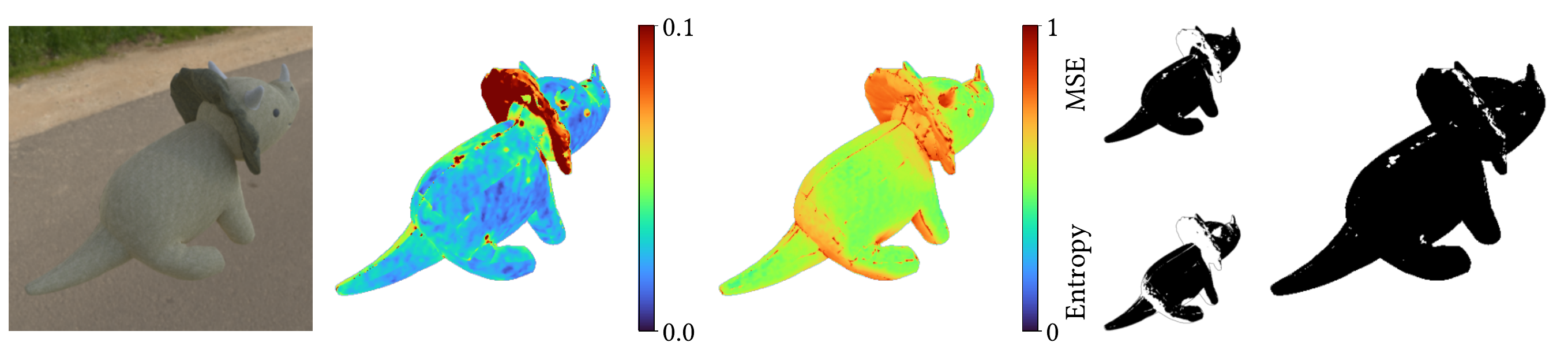}
    \end{subfigure}
    \hspace{0.02\linewidth}
    \begin{subfigure}[b]{0.48\linewidth}
        \centering
        \includegraphics[width=\textwidth]{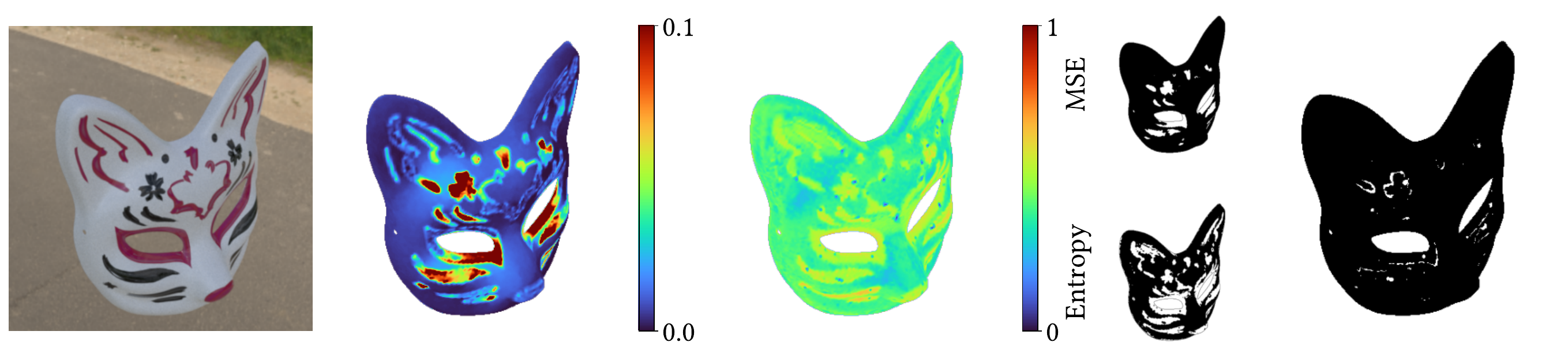}
    \end{subfigure}

    \begin{subfigure}[b]{0.48\linewidth}
        \centering
        \includegraphics[width=\textwidth]{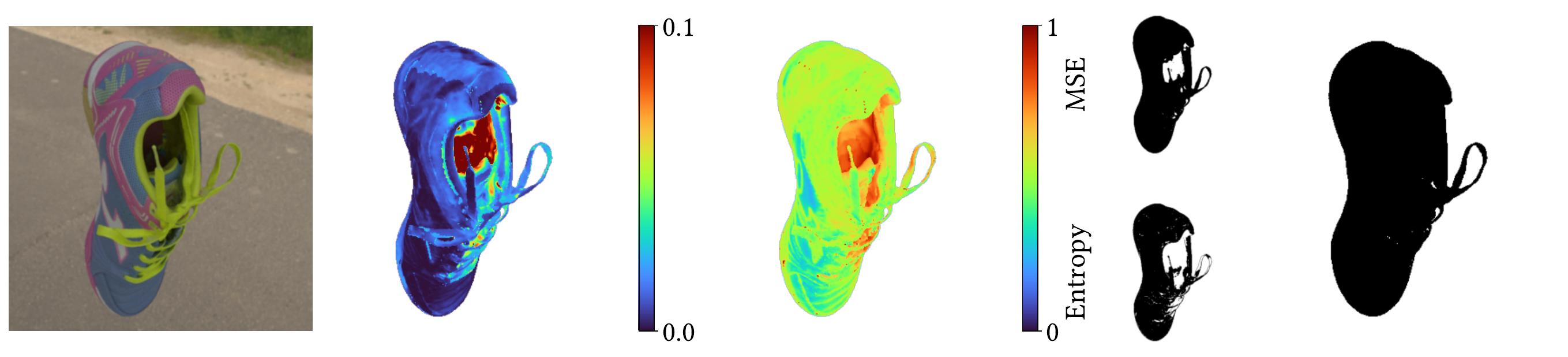}
    \end{subfigure}
    \hspace{0.02\linewidth}
    \begin{subfigure}[b]{0.48\linewidth}
        \centering
        \includegraphics[width=\textwidth]{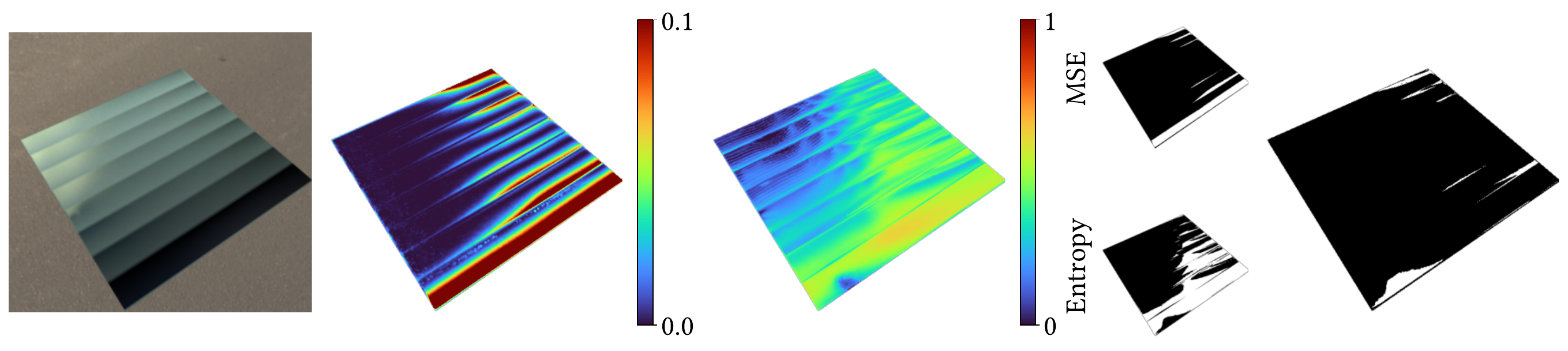}
    \end{subfigure}
    \caption{Comparison of MSE for Mitsuba and entropy computed with the SH Model. We observe that where there is error, entropy tends to be higher. We also show a mask for above-average MSE and entropy. The \textit{Mask Coverage} column shows which regions of the MSE mask are not covered by the entropy mask and should, ideally, be completely black.}
    \label{fig:synthetic_entropy}
    \Description[The figure shows a grid of comparisons on the synthetic benchmark between error (labeled MSE Mitsuba) and Entropy (computed with the SH Model).]{The figure shows a grid of comparisons on the synthetic benchmark between error (labeled MSE Mitsuba) and Entropy (computed with the SH Model). Six objects are shown (dice, donut, dino plushie, cat mask, shoe, plane). Alongside the maps for error and entropy are black-and-white masks that show where the error and entropy are above average. These masks overlap, as shown in a final column, labeled 'Mask Coverage', which is almost fully black for all objects.}
\end{figure*}

%% file: tables/comparisons/uncertainty_stanford_orb.tex
\begin{table}[tb]
\centering
\setlength{\tabcolsep}{3pt}
\scriptsize
\caption{Correlation of entropy computed with Mitsuba and our method.}

\label{tab:benchmark_uncertainty}
\resizebox{0.8\linewidth}{!}{
\begin{tabular}{lcc}
\toprule
 & Entropy Mitsuba $\uparrow$  & Time \\\midrule
    Entropy Mitsuba     &  $1.00 \pm 0.00$  &  $5$m$50$s   \\
    Entropy angular     &  $0.94 \pm 0.04$  &  $1.66$s  \\
 Entropy power spectrum &  $0.88 \pm 0.05$  &  $0.0005$s \\
\bottomrule
\end{tabular}
}
\end{table}

%% file: sections/5_conclusion.tex
\section{Challenges and conclusion}
In summary, we present an uncertainty estimation method for multi-view capture of materials on objects using frequency domain analysis.
We use least-squares fitting and a power spectrum approximation to efficiently compute
entropy as uncertainty. We validate our uncertainty estimation and demonstrate its benefits through multiple applications, improving the quality of captured assets.

A future challenge for practical use is to enable spherical harmonics decompositions on LDR images. Because LDR signals are typically clamped, they introduce frequencies that are not present in the original signal, conflicting with the frequency analysis. Another direction is extending the framework for situations with unknown lighting, which requires computing entropy over hypothetical lighting setups. We believe frequency analysis for uncertainty estimation provides a powerful and practical tool for understanding and improving object capture approaches and look forward to its applications in future work.

%% file: sections/6_acknowledgements.tex
\begin{acks}
We would like to thank Yash Belhe and Georgios Kopanas for insightful early discussions. This work was partially supported by the convergence theme AI, Data and Digitalisation of Erasmus University, the Erasmus medical centre and the Delft University of Technology via the Immersive Tech AI lab.
\end{acks}

%% file: sections/A_appendix.tex
\section{Details on- and Extensions to the Convolution Model}
In this section, we provide a derivation of the convolution model for reflection. This derivation does not add technical novelty, but provides clarity on certain details that were implied in the original paper by \citeauthor{Ramamoorthi_Hanrahan_2001}.
We also propose a simple extension to the convolution model to include shadowing and masking effects, which were ignored by \citeauthor{Ramamoorthi_Hanrahan_2001}.
Finally, we discuss how to map the parameters of the Torrance-Sparrow BRDF to the Disney Principled BRDF \cite{burley2012physically}, which is commonly used in modern rendering pipelines.

\subsection{Derivation of the Convolution Model}
\label{sec:derivation}
The outgoing radiance at point $p$ along direction $\omega_o$, $L_o(p, \omega_o)$ is given by
\begin{equation}
B(p, \omega_o)=\int_{H^2(\mathbf{n})}f_r(p, \omega_o, \omega_i)L(p, \omega_i)\cos\theta_id\omega_i,
\label{eq:reflection}
\end{equation}
where $f$ is the BRDF and $L(p, \omega_i)$ is the incident radiance along direction $\omega_i$. For the Torrance-Sparrow BRDF, $f$ is defined as
\begin{equation}
f(p, \omega_o, \omega_i) = K_d + K_s\frac{D(\omega_m)F(\omega_o\cdot\omega_m)G(\omega_i, \omega_o)}{4\cos\theta_i\cos\theta_o},
\label{eq:torrancesparrow_full}
\end{equation}
where $\omega_m$ is the half-direction vector $\omega_m = (\omega_i + \omega_o) / ||\omega_i + \omega_o||$;
$D(\omega_m)$ is the normal distribution function;
$F(\omega_o\cdot\omega_m)$ is the Fresnel term. Ramamoorthi and Hanrahan simplify this term to $F(\theta_o)$, as the angle $\theta_o$ is often close to the angle between $\omega_o$ and $\omega_m$;
$G(\omega_i, \omega_o)$ is the shadowing-masking term. \citeauthor{Ramamoorthi_Hanrahan_2001} ignore $G$. We assume shadowing and masking are independent statistical events, so that $G(\omega_i, \omega_o)=G(\omega_i)G(\omega_o)$.

There are two important notes about the denominator in \autoref{eq:torrancesparrow_full}:
\begin{enumerate}
    \item $1/(4\cos\theta_o)$ results from the half-direction transform: the distribution of microfacets with a normal $\omega_m$ is transformed to the distribution of outgoing directions $\omega_o$ that the incoming light ray $\omega_i$ reflects toward (see \citeauthor{pharr2023physically}, Equation 9.27).
    \item $1/(\cos\theta_i)$ cancels out the cosine term applied to the incoming radiance (see \citeauthor{pharr2023physically}, equation 9.30).
\end{enumerate}

We now substitute \autoref{eq:torrancesparrow_full} into \autoref{eq:reflection} and split the equation into diffuse and specular
\begin{align}
B(p, \omega_o)
&=K_d \int_{H^2(\mathbf{n})} L(p, \omega_i)\cos\theta_id\omega_i \\
&+ K_s\int_{H^2(\mathbf{N})} \frac{D(\omega_m)F(\omega_o\cdot\omega_m)G(\omega_i, \omega_o)}{4\cos\theta_i\cos\theta_o} L(p, \omega_i)\cos\theta_id\omega_i\notag
\end{align}
This equation is simplified by \citeauthor{Ramamoorthi_Hanrahan_2001} using the assumptions that $F$ only depends on $\theta_o$ and shadowing-masking is ignored. We replace the integral of incoming radiance for diffuse with the symbol for irradiance $E$.
\begin{align}
B(p, \omega_o)&=K_d E(p) + K_s F(\theta_o)\int_{H^2(\mathbf{N})} \frac{D(\omega_m)}{4\cos\theta_o} L(p, \omega_i)d\omega_i.
\label{eq:reflection_simplified1}
\end{align}
\citeauthor{Ramamoorthi_Hanrahan_2001} rewrite the specular term as a convolution between a filter based on $D$, and $L$. Crucially, the domain of $D$ in the Torrance-Sparrow model is the half-angle space. In \citeauthor{Ramamoorthi_Hanrahan_2001}'s derivation, the spherical harmonic representation for this filter, in the paper referred to as $S$, is derived in incoming-direction space for normal exitance (\citeauthor{Ramamoorthi_Hanrahan_2001}, Equation 27). This has two consequences:
\begin{enumerate}
    \item We do not have to account for a change of variables and $1/(4\cos\theta_o)$ can be removed.
    \item In reality, $S$ depends on the outgoing direction that is observed and thus, the filter changes shape. This variation is ignored with the explanation that ``the BRDF filter is essentially symmetric about the reflected direction for small viewing angles, as well as for low frequencies $l$. Hence, it can be shown by Taylor-series expansions (and verified numerically) that the corrections to Equation 20 [\autoref{eq:microfacet_brdf} in our paper] are small under these conditions.''
\end{enumerate}
This means that we can rewrite \autoref{eq:reflection_simplified1} with a convolution
\begin{equation}
B(p, \omega_o)=K_d E(p) + K_s F(\theta_o)\left[S \ast L \right]_{\omega_o},
\label{eq:reflection_simplified2}
\end{equation}
which equals Equations 21 and 22 in \citeauthor{Ramamoorthi_Hanrahan_2001}.

\begin{figure*}[t]
    \centering
    \includegraphics[width=\textwidth]{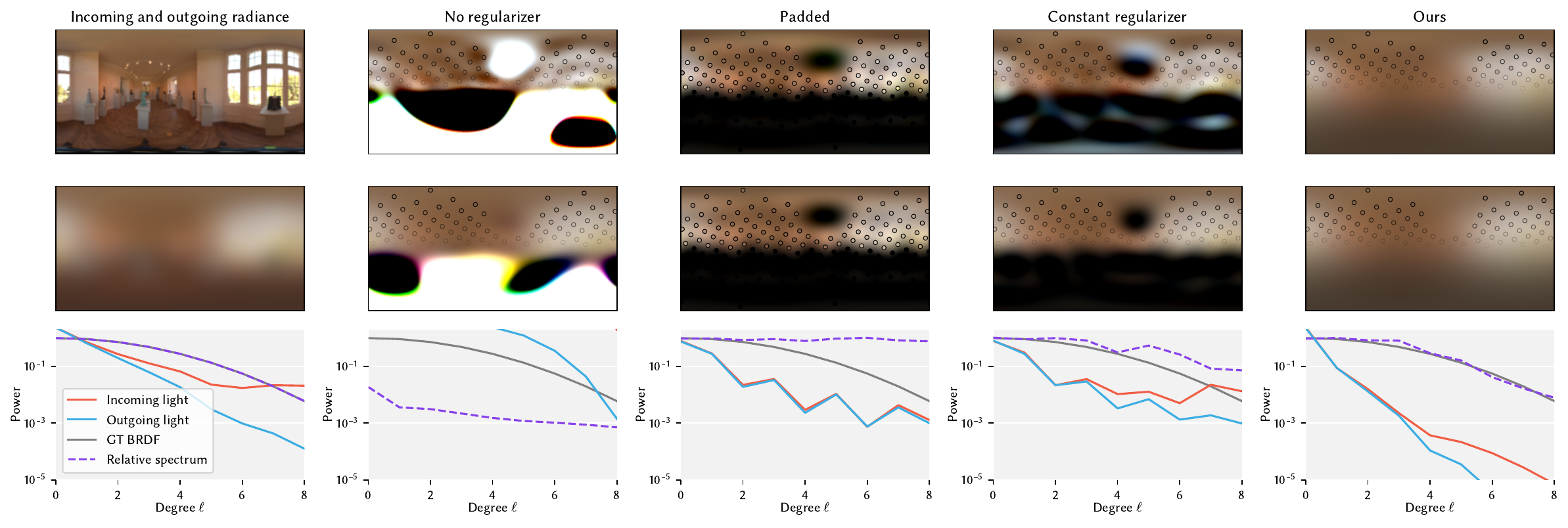}
    \caption{A comparison of the options for fitting spherical harmonics on sparse samples. In the first column, top: incoming radiance (synthetic) on the sphere mapped to a lat-long grid. Middle: synthesized outgoing radiance as a result of filtering incoming radiance with \autoref{eq:microfacet_brdf}, $\alpha=0.2$. Bottom: power spectra of incoming and outgoing radiance and ground-truth filter (BRDF) plotted on a log-scale. In the second to fifth column, we show the result of fitting spherical harmonics coefficients to samples and then transforming back to the spatial domain. We use $100$ samples from the upper hemisphere of the input radiance (first column) and simulate missing samples due to occlusion or missed camera positions by masking some points, leaving $88$ samples. The samples are weighted with $\cos\theta$, which is visualized as the transparency of the samples. We find that our method with a weighted regularizer is able to smoothly interpolate missing values and retrieves the correct BRDF filter in the power spectrum, where other variants overfit, or add dark regions in the upper hemisphere.}
    \label{fig:sh_lsq}
\end{figure*}

\subsection{Shadowing and Masking} The approximate reflection function in \autoref{eq:reflection_simplified2} does not include the shadowing or masking terms present in microfacet models~\cite{pharr2023physically}.
These terms model occlusions of incoming light (shadowing) and outgoing light (masking) due to the configuration of the microfacets.
\citeauthor{Ramamoorthi_Hanrahan_2001} argue that these terms can be ignored, because they mostly affect observations made at grazing angles.
While this is true for materials with low roughness, we find that ignoring this term can lead to an appearance mismatch for high-roughness materials.
We propose a simple way to include these terms in the convolution model to study the effect on uncertainty and BRDF acquisition.
Shadowing and masking effects are typically modeled jointly to avoid an over-correction of the radiance (\citet{pharr2023physically}, Section 9.6.3). However, this joint term breaks the convolution model as presented in \autoref{eq:reflection_simplified2}, as the kernel would depend on both $\omega_i$ and $\omega_o$. 
Therefore, we assume that shadowing and masking are independent and model them as $G_\alpha(\omega_i)G_\alpha(\omega_o)$.
We first attenuate the incoming light with the shadowing term, convolve with the BRDF and attenuate the result with the masking term:
\begin{equation}
    B(p, \omega_o)\approx K_d E(p) + K_s F(\theta_o) G_\alpha(\theta_o)\left[S_\alpha \ast G_\alpha(\theta_i) L(p) \right]_{\omega_o},
    \label{eq:shadowing_masking_conv_framework}
\end{equation}
where $G_\alpha$ is the shadowing-masking function in the \citeauthor{trowbridge1975average} model.

\subsection{Torrance-Sparrow to Principled BRDF}
Many modern rendering pipelines employ variants of the Disney BRDF~\cite{burley2012physically}, which is a combination of a diffuse term and a microfacet term with a user-friendly parametrization. The model also contains some additional features beyond the scope of the current work. We can formulate our model using the principled BRDF parameters, rather than the raw parameters of the Torrance-Sparrow model. We parameterize base color, metallicity, and roughness, mapping these to the Torrance-Sparrow model as
\begin{align}
    K_d &= R_b,\\
    K_s &= 1,\\
    R_0 &= 0.04 + (R_b - 0.04) m,\\
    F(\theta_o) &= R_0 + (1 - R_0) (1 - \cos\theta_o)^5,\\
    \alpha &= r^2,
\end{align}
where $R_b$ is the base color, $m$ is metallicity and $r$ is the roughness. We set $K_s$ to $1$, because the specular component is already scaled in the Fresnel term by $R_0$. This parameterization is based on Schlick's approximation~\cite{schlick1994} and follows the implementation in Mitsuba~\cite{jakob2022mitsuba3}.

\section{Least-Squares Regularization}
We experiment with a number of options for least-squares fitting and show the results in \autoref{fig:sh_lsq}. In these experiments, we found that no regularization and constant regularization lead to incorrect estimations of the BRDF, which can be retrieved as the ratio of the outgoing radiance over the incoming radiance power spectrum.
Our proposed solution using a non-constant regularizer leads to a smooth interpolation of missing regions. While it also `blurs' the incoming radiance, the relative relation between incoming and outgoing radiance is preserved.

\section{Sampling Theory}
The transformation from the directional domain to spherical harmonics begs the question: do we have enough samples to accurately recover the coefficients of the outgoing radiance? We know from \autoref{eq:microfacet_brdf} that the BRDF acts as a low-pass filter parameterized by $\alpha$. We connect this knowledge with sampling theory to derive lower bounds on sampling counts.

The Nyquist-Shannon theorem provides a lower bound on the number of samples required to exactly recover a band-limited signal using a Fourier series. Similar theorems have been developed for spherical harmonics~\cite{mcewen2011compressedSH, mcewen2011novel, DRISCOLL1994202}. These state that, to recover a spherical signal with band-limit $\ell^{*}$, the number of samples should be $\mathcal{O}(\ell^{*2})$.
The sampling rate and related band-limit have direct consequences for BRDF recovery. Assume that the incoming light has been sampled at a high enough rate to be accurately recovered, for example, from projected photographs or a gazing sphere. Then the outgoing light is the weakest link, as it is sampled by moving the camera along $N$ positions around the object. Sampling theory tells us that we can only accurately recover outgoing radiance that is band-limited to $\ell^{*} < \sqrt{N}$ degrees. Signals with non-zero amplitude in higher degrees will suffer from aliasing.

Fortunately, the BRDF acts as a low-pass filter on the incoming radiance (\autoref{eq:microfacet_brdf}). That means the outgoing radiance can fall into two categories, based on the $\alpha$ parameter of the material ($\alpha = \text{roughness}^2$): $\alpha$ is either too low or $\alpha$ is high enough to recover spherical harmonic coefficients. If $\alpha$ is too low, the low-pass filtering from the BRDF does not band-limit the signal enough to accurately recover with the given sampling rate. The threshold for $\alpha$ can be determined based on \autoref{eq:microfacet_brdf}. Let $t$ be an acceptable attenuation factor for degrees $\ell > \ell^{*}$. We solve \autoref{eq:microfacet_brdf} for $t$ to find the lower bound, $\alpha'$, for accurate recovery
\begin{equation}
   \alpha' = \ell^{*-1} \sqrt{-\ln t}.
\end{equation}
An acceptable threshold $t$ can be determined empirically, by investigating the reconstruction error for a set of environment maps. To provide some intuition, for $N=400$ samples and a threshold of $t=0.5$, $\alpha' \approx 0.07$. Above this threshold, our method can recover $\alpha$ and $K_s$, provided that the incoming radiance has enough amplitude in the right degrees. This also extends to non-uniform samples, because the Nyquist-Shannon theorem holds for non-uniform samples~\cite{marvasti2012nonuniform}. In other words: if a lower bound on $\alpha$ is known, it does not matter where the camera is placed, as long as the average distance to the closest sample is equal to $1/N$. It also means that one can determine the number of required views based on the lowest $\alpha$ that should be recovered: $N \sim \alpha^{-2}$.

It is important to understand what happens if $\alpha < \alpha'$. First, we would be uncertain where $\alpha$ lands between $0$ and $\alpha'$, based on the power spectrum alone. For $0 < \alpha < \alpha'$, \autoref{eq:microfacet_brdf} is close to $1$ for all degrees below $\ell^{*}$. Second, because this situation occurs for low $\alpha$, the outgoing radiance should be similar to the incoming radiance, up to a scale factor for absorption and transmission. It is unlikely that the spherical harmonics decomposition with significant aliasing will match a filtered version of the incoming radiance. Therefore, we can detect that $\alpha < \alpha'$. In this case, the MSE for any parameter combination $\psi$ is relatively high. Once such a case is detected, we know that our spherical harmonic-based analysis provides no further insights on (un)certainty. There is still a chance for accurate BRDF recovery if $\alpha < \alpha'$. A sample might land on a fortunate spot in the outgoing radiance field. This is the case when there is high local variation in the incoming light around the sample locations, resulting in large changes in radiance for small changes in $\alpha$. 

\section{Background}
Our method builds on prior work in inverse rendering and spherical harmonics. We summarize required background knowledge and refer to related work for further depth.

\subsection{\emph{Reflection as Convolution}}
In the inverse rendering framework of \citet{Ramamoorthi_Hanrahan_2001}, detailed in \autoref{sec:derivation}, estimating the specular BRDF parameters comes down to estimating the convolution kernel $S$.
This can be done efficiently since a convolution in the angular domain can be represented as a multiplication in the spherical harmonics frequency domain. Using this representation, one can find the convolution kernel through a division of the spherical harmonics coefficients of the outgoing radiance by those of the incoming radiance. This is analogous to kernel estimation for image deblurring in the Fourier domain.
\citeauthor{Ramamoorthi_Hanrahan_2001} derive a number of conclusions from this insight, which we summarize in this section.

One crucial insight concerns the well-posedness of BRDF recovery. \citeauthor{Ramamoorthi_Hanrahan_2001} state that the recovery of BRDF parameters is ill-posed if the input lighting has no amplitude along certain modes of the filter (BRDF). Those modes cannot be estimated without additional priors on plausible spatial parameter variations. For the microfacet BRDF, this leads to the following conclusion: if the incoming light only contains frequencies $\ell << \alpha^{-1}$, multiplying the coefficients of the light with those of the BRDF only results in a small difference, and the inversion of this operation is ill-conditioned. To accurately estimate $\alpha$, the incoming light used during the capture needs to exhibit sufficiently high frequencies.

This insight is based on a derivation for the coefficients of the microfacet model. The normalized SH-coefficients of the specular component of the BRDF for normal incidence, $S$, are approximated by
\begin{equation}
\hat f_{\ell m} \approx e^{-(\alpha \ell)^2},
\label{eq:microfacet_brdf}
\end{equation}
which is a Gaussian in the frequency domain with a width determined by $\alpha$. The kernel is derived from a Beckmann normal distribution function and the $\alpha$ parameter corresponds to the $\alpha$ parameter there. Note that these coefficients do not vary with the Spherical Harmonics order $m$, since the normal distribution function is isotropic for outgoing rays in the direction of the normal vector ($\theta_o = 0$). An important approximation employed by \citeauthor{Ramamoorthi_Hanrahan_2001} is that this same kernel can be used for any outgoing direction. While the correct kernel varies with the incoming and outgoing direction, this approximation does not lead to significant error for inverse rendering~\cite{Ramamoorthi_Hanrahan_2001} as they note that \emph{``it can be shown by Taylor-series expansions and verified numerically, that the corrections to this filter are small} [for low degrees $\ell$]''.

In this paper, we expand on the theory established by \citeauthor{Ramamoorthi_Hanrahan_2001} by improving the reflection as a convolution model's accuracy and developing the implications for well-posedness into quantifiable metrics on uncertainty without a-priori knowledge on $\alpha$.

\subsection{Spherical Harmonics}
Spherical harmonics are a series of orthonormal basis functions on the sphere, indexed by their degree $\ell$ and order $m$. We use them to represent incoming and outgoing radiance over incoming and outgoing directions on the unit sphere. Spherical harmonics are analogous to the Fourier series on a flat domain, where the frequency of the Fourier series corresponds to the degree $\ell$ and order $m$. We provide a brief overview of properties relevant to our method. For further details, a helpful reference and software package is published by \citet{wieczorek2018shtools}.

Any real, square-integrable function on the sphere can be expressed as a spherical-harmonics series:
\begin{equation}
f(\theta, \phi) = \sum_{\ell=0}^{\infty}\sum_{m=-\ell}^{\ell}f_{\ell m}Y_{\ell m}(\theta, \phi),
\label{eq:sh_transform}
\end{equation}
where $f_{\ell m}$ is the coefficient for spherical harmonic $Y_{\ell m}(\theta, \phi)$, given as
\begin{align}
N_{\ell m} &= \sqrt{\frac{(2 - \delta_{m0})(2\ell + 1)}{4\pi}\frac{(\ell-m)!}{(\ell+m)!}}\\
Y_{\ell m}(\theta, \phi) &= \begin{cases}
    N_{\ell m}P_\ell^m(\cos\theta)\cos m\phi & \text{if $m \geq 0$,}\\
    N_{\ell m}P_\ell^{|m|}(\cos\theta)\sin |m|\phi & \text{if $m < 0$.}
\end{cases}
\label{eq:sphericalharmonics}
\end{align}
$N_{\ell m}$ is a normalization factor, $\delta_{m0}$ is the Kronecker delta function, which evaluates to $1$ when $m=0$, and $P_\ell^m$ is the associated Legendre function for degree $\ell$ and order $m$. The total number of spherical harmonics up to- and including a maximum degree, $\ell^{*}$, equals $(\ell^{*}+1)^2$. A useful property of spherical harmonics in our setting is that a rotational convolution on the sphere is equal to multiplication of coefficients in the spherical harmonic domain. 

The power spectrum of a spherical function $f$ can be computed from the spherical harmonic coefficients per degree
\begin{equation}
S_{f}(\ell)=\sum_{m=-\ell}^{\ell}f_{\ell m}^2.
\end{equation}
The power spectrum is invariant to rotations of the coordinate system. In our context that means the power spectrum is invariant to slight perturbations of the normals at each point.

\subsubsection{Computing spherical harmonic coefficients}
One can find the SH coefficients for a function $f$ by computing the inner product with the basis functions
\begin{equation}
f_{\ell m} = \int_{S^2} f(\theta, \phi)Y_{\ell m}(\theta, \phi)d\omega.
\label{eq:fullshtransform}
\end{equation}
A useful property holds for the coefficient of degree $\ell=0$, for which the spherical harmonic is constant; $Y_{00}(\theta, \phi) = (4\pi)^{-\frac{1}{2}}$. The corresponding coefficient, $f_{00}$, is equal to the integral of $f$ times the normalization constant, $(4\pi)^{-\frac{1}{2}}$. The spherical harmonics for higher degrees all integrate to zero\footnote{Because the spherical harmonics are orthormal, the inner product between any spherical harmonic with $\ell > 0$ and the constant function ($\ell, m = 0$) equals zero.}. This is relevant in the context of rendering, because the total integrated incoming and outgoing radiance can be read from the $0^{th}$ degree coefficient and that coefficient alone. In the general case, we estimate the coefficient for $f_{\ell m}$ based on samples of $f$. The sampling method determines how these coefficients are estimated.

\paragraph{Regular sampling} If $f$ is sampled on a grid with equally spaced longitudinal and latitudinal angles, this integral can be accelerated using a fast Fourier transform in the longitudinal direction $\phi$ and a quadrature rule in the latitudinal direction $\theta$~\cite{DRISCOLL1994202}. In our setting, this approach can be used for environment maps represented as rectangular textures.

\paragraph{Irregular sampling} During capture the camera is often placed at irregular positions, leading to non-uniform $(\theta,\phi)$ samples. 
Further, a point on the surface might be observed from only a few positions.
We therefore often need to use sparse and irregular samples to fit spherical harmonic coefficients.
We do so by fitting the coefficients using least-squares, expressing \autoref{eq:sh_transform} as a linear system
\begin{equation}
\mathbf{Y}\mathbf{c} \approx \mathbf{f},
\label{eq:sh_linear}
\end{equation}
where $\mathbf{f}$ is a vector of $n$ discrete samples from $f$, $\mathbf{Y}$ is a matrix of size $n \times (\ell^{*} + 1)^2$ containing the spherical harmonics sampled at the same locations as $\mathbf{f}$, and $\mathbf{c}$ is a vector of the $(\ell^{*} + 1)^2$ coefficients we want to find. We can find $\mathbf{c}$ by solving a least-squares system \begin{equation}
\mathbf{Y}^\intercal\mathbf{Y}\mathbf{c} = \mathbf{Y}^\intercal \mathbf{f}.
\label{eq:sh_lsq}
\end{equation}
To be well posed, this system requires $n > (\ell^{*} + 1)^2$ independent samples, which can be challenging in the context of sparse sampling, making the system under-constrained.
We propose to use a custom regularizer in our work for cases where the number of samples is too low.

\section{Stanford ORB Reference Results}
We include the results table from StanfordORB for reference. These results were obtained under different acquisition condition and cannot be directly compared to our results.
\input{tables/comparisons/stanford_orb_others}

%% file: tables/comparisons/stanford_orb_others.tex
\begin{table}[t]
\centering
\setlength{\tabcolsep}{3pt}
\scriptsize
\caption{Benchmark Comparison for \textbf{Novel Scene Relighting} of Existing Methods from \cite{kuang2023stanfordorb}. \textdagger~denotes models trained with the ground-truth 3D scans and pseudo materials optimized from light-box captures. The rest of results are obtained by optimizing jointly for illumination, geometry and material. \emph{We report these numbers for reference, however they cannot be directly compared to our results.}}

\label{tab:benchmark_stanfordorb}
\resizebox{1.0\linewidth}{!}{
\begin{tabular}{lcccc}
\toprule
&  PSNR-H$\uparrow$ & PSNR-L$\uparrow$ & SSIM$\uparrow$ & LPIPS$\downarrow$  \\\midrule
NVDiffRecMC~\cite{hasselgren2022shape}~\textdagger & $25.08$ & $32.28$ & $0.974$ & $0.027$  \\
NVDiffRec~\cite{munkberg2022extracting}~\textdagger & $24.93$ & $32.42$ & $0.975$ & $0.027$  \\
\midrule
PhySG~\cite{zhang2021physg} & $21.81$ & $28.11$ & $0.960$ & $0.055$  \\
NVDiffRec~\cite{munkberg2022extracting} & $22.91$ & $29.72$ & $0.963$ & $0.039$  \\
NeRD~\cite{boss2021nerd} & $23.29$ & $29.65$ & $0.957$ & $0.059$  \\
NeRFactor~\cite{zhang2021nerfactor} & $23.54$ & $30.38$ & $0.969$ & $0.048$  \\
InvRender~\cite{wu2023nefii} & $23.76$ & $30.83$ & $0.970$ & $0.046$  \\
NVDiffRecMC~\cite{hasselgren2022shape} & $24.43$ & $31.60$ & $0.972$ & $0.036$  \\
Neural-PBIR~\cite{sun2023neural} & $26.01$ & $33.26$ & $\mathbf{0.979}$ & $\mathbf{0.023}$  \\

\bottomrule
\end{tabular}
}
\end{table}